\RequirePackage{fix-cm}

\documentclass[twocolumn]{svjour3}

\smartqed 

\usepackage{natbib}
\usepackage{graphicx}
\usepackage{subcaption}
\usepackage{multirow}
\usepackage{booktabs}
\usepackage{amsmath,amssymb}
\usepackage{algorithm}
\usepackage{algorithmic}
\usepackage{arydshln}
\usepackage{caption}
\usepackage{colortbl}
\usepackage[dvipsnames]{xcolor}
\usepackage{color, colortbl}
\definecolor{citecolor}{HTML}{0071bc}
\definecolor{tabhighlight}{HTML}{e5e5e5}
\usepackage[colorlinks,citecolor=citecolor]{hyperref}
\usepackage{microtype}

\begin{document}

\title{WiT: Waypoint Diffusion Transformers for Alleviating Trajectory Conflict in Pixel-Space Image Generation
}

\author{Hainuo Wang* \and
        Mingjia Li* \and
        Xiaojie Guo\dag
}

\institute{Hainuo Wang \at
              School of Computer Science and Technology, Tianjin University, Tianjin, China \\
              \email{hainuo@tju.edu.cn}
           \and
           Mingjia Li \at
              School of Software, Tianjin University, Tianjin, China \\
              \email{mingjiali@tju.edu.cn}
           \and
           Xiaojie Guo  \at
              School of Software, Tianjin University, Tianjin, China \\
              \email{xj.max.guo@gmail.com}
           \and
           * Equal Contribution.
           \dag  \,Corresponding author.
}

\date{Received: date / Accepted: date}

\maketitle

\begin{abstract}
While recent Flow Matching models avoid the reconstruction bottlenecks of latent autoencoders by operating directly in pixel space, the raw pixel manifold provides little explicit semantic organization, making target-specific transport directions difficult for a shared finite-capacity vector field to distinguish. When different transport requirements are locally entangled in this way, their learning signals can interfere and hinder optimization, a phenomenon we refer to as \textit{trajectory conflict}. To alleviate trajectory conflict while preserving direct generation in pixel space, we propose \textbf{W}aypoint D\textbf{i}ffusion \textbf{T}ransformers (WiT), which introduces explicit semantic routing into pixel-space generation. WiT structures pixel-space prediction through compact semantic waypoints projected from pre-trained vision representations, providing a semantically organized intermediate representation that complements direct pixel prediction. During ODE integration, a lightweight Waypoints Predictor dynamically infers the semantic waypoint from the current noisy state, and the predicted waypoint continuously conditions the primary Pixel Space Generator through our Just-Pixel AdaLN mechanism. This spatially varying semantic modulation reorganizes the finite-capacity learning problem and helps the model preserve target-specific transport directions while retaining generation entirely in pixel space. Experiments on ImageNet 2012 demonstrate consistent improvements over strong pixel-space baselines across model scales, with WiT-H/16 achieving an FID of 1.79. Code will be released on our \href{https://hainuo-wang.github.io/WiT}{project page}.
\end{abstract}

\begin{figure*}[t]
    \centering
    \includegraphics[width=0.95\linewidth]{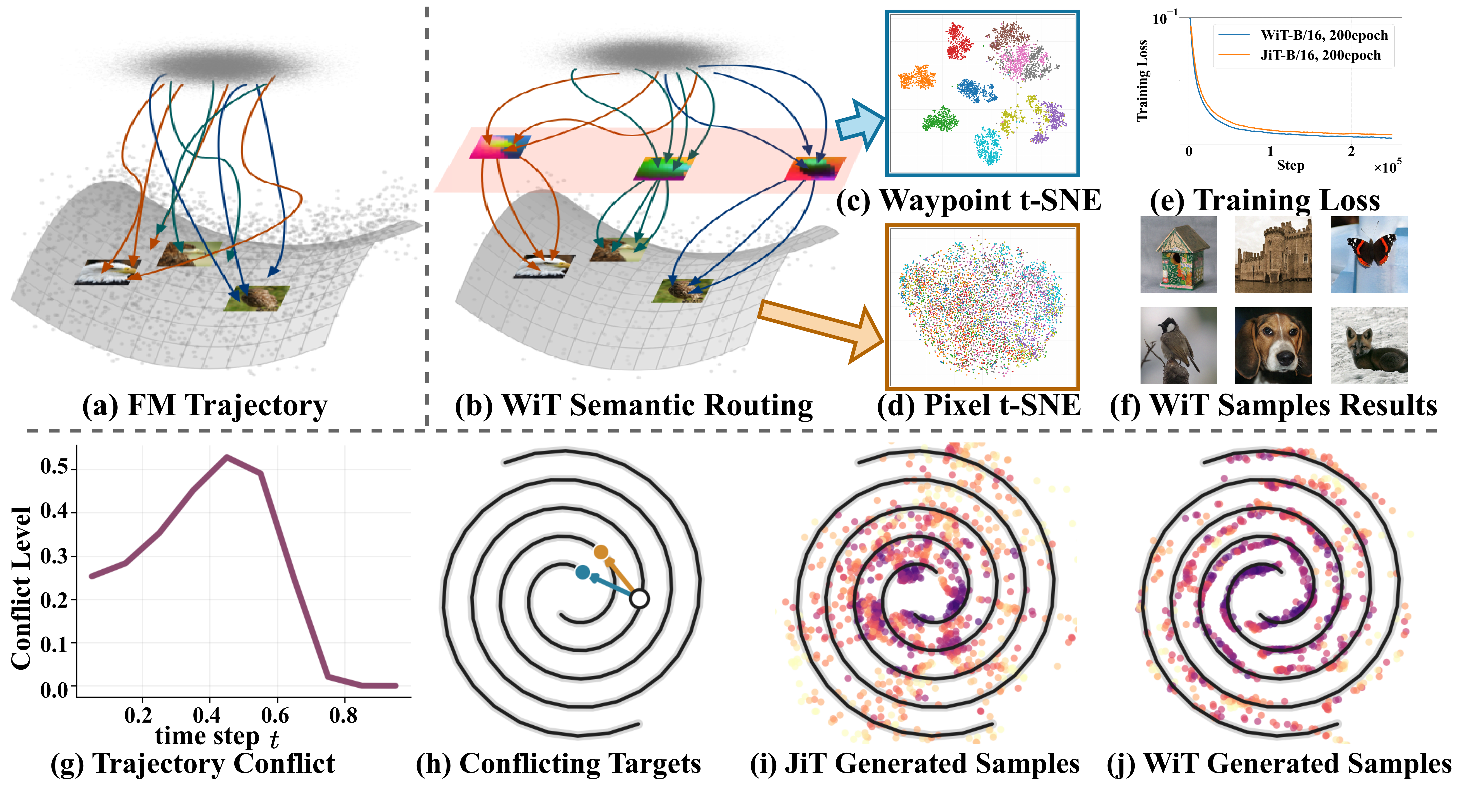}
    \caption{Overview and toy illustration of WiT. (a)--(b) compare standard pixel-space prediction with semantic routing, while (c)--(d) contrast waypoint and raw-pixel representations. (e)--(f) show training-loss curves and representative samples. (g) plots the intrinsic semantic-conflict ratio on the analytic two-spiral distribution, computed as the between-semantic-region posterior variation divided by the total endpoint posterior variation, with the strongest conflict occurring at intermediate timesteps. (h) illustrates a representative local conflict, where nearby noisy states are associated with different target directions. (i)--(j) compare JiT and WiT on the two-spiral toy experiment, using two-hidden-layer MLPs with width 128 in a 64-dimensional observation space. Samples closer to the target manifold indicate better recovery.}
    \label{fig:motivation}
\end{figure*}

\section{Introduction}
\label{sec:intro}

Diffusion models~\citep{Ho2020, Song2021}, particularly those formalized through Flow Matching (FM) frameworks~\citep{lipman2023flowmatchinggenerativemodeling, liu2022flow, albergo2023building} and scaled via Diffusion Transformers (DiT)~\citep{Peebles2023, ma2024sit}, have established a new standard in highly realistic image generation. To improve computational efficiency and facilitate stable optimization and convergence, these architectures commonly operate in latent spaces~\citep{shi2025latent, Rombach2022, blattmann2023stable}, relying on continuous-valued variational autoencoders (VAEs)~\citep{Rombach2022, Esser2021, Mentzer2024, wang2026ddt} to compress raw visual signals into lower-dimensional representations. However, this two-stage design also introduces an inherent information bottleneck. The visual tokenizer must compress a high-dimensional image into a substantially lower-dimensional latent representation, and this compression cannot preserve all information contained in the original pixel signal~\citep{Esser2021,Rombach2022}. Fine-grained details such as local textures, sharp boundaries, and subtle spatial variations may therefore be weakened or discarded during encoding~\citep{Rombach2022}. Since the downstream generative model learns to produce samples only within this compressed representation space, information that is absent from the latent representation cannot be explicitly modeled by the generative process and must instead be reconstructed by the decoder. As a result, reconstruction errors and tokenizer-induced artifacts can propagate to the final generated images, potentially limiting the achievable pixel-level quality~\citep{yao2025reconstruction}. 

To overcome the reconstruction bottleneck of latent generative pipelines, recent work, exemplified by JiT~\citep{Li_2026_CVPR}, advocates learning continuous vector fields directly in pixel space~\citep{Yu_2026_CVPR, ma2026deco, Chen_2026_CVPR, lei_there_2026}. By bypassing the visual tokenizer, pixel-space Flow Matching avoids compression-induced reconstruction errors and directly models fine-grained image details. This benefit, however, comes with a harder optimization problem, as a finite-capacity generator must learn high-dimensional pixel-space structure directly from noisy RGB observations~\citep{yao2025reconstruction, bfl2025representation}. Meanwhile, recent latent-space studies such as VA-VAE~\citep{yao2025reconstruction} show that aligning generative representations with pre-trained vision features can provide more semantically organized representations and facilitate optimization. Pure pixel-space generation, however, must preserve raw RGB as the generative target, making representation-alignment strategies that reshape the target latent space inapplicable without changing the underlying modeling domain. As illustrated in Figure~\ref{fig:motivation}(c)--(d), this motivates us to introduce explicit semantic organization into the prediction process while retaining raw pixels as the generative target.

This lack of explicit semantic organization becomes particularly important when the noisy state only partially reveals the target structure. Along the flow trajectory, noise obscures target-specific cues such as pose, layout, scale, and spatial arrangement, while local proximity in noisy RGB space does not necessarily imply similarity in the underlying target structure. As illustrated in Figure~\ref{fig:motivation}(g)--(h), nearby noisy states under the same external condition can therefore be associated with different target-specific structures and transport requirements. Under the Flow Matching~\citep{liu2022flow, lipman2023flowmatchinggenerativemodeling}, a shared finite-capacity vector-field model must resolve these locally mixed requirements, which can produce interfering learning signals and under-resolved transport directions. We refer to this optimization difficulty as \textit{trajectory conflict}.
Although Classifier-Free Guidance (CFG)~\citep{ho2022classifier} can strengthen external conditioning at inference time, it does not provide the spatially structured semantic organization needed during training. This motivates our central question:
\textit{How can we introduce spatially discriminative semantic structure into pixel-space Flow Matching while retaining direct generation in the raw pixel domain?}

To address this problem, we introduce an explicit semantic waypoint into pixel-space Flow Matching. Rather than reshaping the raw RGB target space or decomposing generation into separate sequential flows, WiT structures the prediction problem through a compact intermediate semantic representation. A lightweight Waypoints Predictor dynamically predicts a semantic waypoint from the current noisy pixel state, capturing spatial structure that is then exposed to the primary Pixel Space Generator as a routing signal. In this way, the generative trajectory remains entirely in pixel space, while the Pixel Space Generator receives semantically organized guidance that would otherwise need to be inferred implicitly from noisy RGB observations. As illustrated in Figure~\ref{fig:motivation}(a)--(d), this semantic routing introduces a more discriminative intermediate representation for organizing target-specific transport requirements.

We instantiate this idea with WiT (\textbf{W}aypoint D\textbf{i}ffusion \textbf{T}ransformers), a framework designed to alleviate trajectory conflict in pixel-space Flow Matching. First, we project frozen vision representations into a compact semantic subspace using Principal Component Analysis (PCA), substantially reducing the dimensionality of the waypoint target while retaining its dominant semantic structure. Second, a lightweight Waypoints Predictor dynamically predicts this compact representation from the current noisy pixel state at each integration timestep $t$. Third, the primary Pixel Space Generator is conditioned on the predicted waypoint through our proposed Just-Pixel AdaLN mechanism, which converts the semantic representation into spatially varying modulation parameters throughout the transformer. As the pixel state evolves, the waypoint is continuously updated, providing an adaptive semantic routing signal for generation.

This design introduces semantic organization into pixel-space prediction without reintroducing a latent image bottleneck or moving the generative state outside the raw pixel domain. Figure~\ref{fig:motivation}(e)--(j) provides complementary illustrations of training behavior, manifold recovery in the toy example, and generated samples. Our analyses further show that the predicted waypoint induces neighborhoods with more coherent semantic endpoints and reduces disagreement-dependent output-layer gradient interference. Experiments on ImageNet 2012~\citep{deng2009imagenet} demonstrate consistent improvements over the direct pixel-space baseline JiT~\citep{Li_2026_CVPR} across model scales and training budgets. Our main contributions are summarized as follows:
\begin{itemize}
    \item We propose Waypoint Diffusion Transformers (WiT), a pixel-space generative framework that introduces explicit semantic routing to alleviate trajectory conflict. WiT structures the finite-capacity prediction problem through dynamically predicted semantic waypoints while keeping the generative trajectory entirely in pixel space.
    
    \item We introduce Just-Pixel AdaLN, which converts semantic waypoints into spatially varying affine modulation for the Pixel Space Generator, providing dense spatial semantic guidance without extending or replacing the native pixel-token sequence.
    
    \item We provide geometric and optimization-level analyses of trajectory conflict, showing that semantic routing reorganizes local semantic neighborhoods and reduces disagreement-dependent gradient interference. Experiments on ImageNet 2012 further demonstrate consistent improvements over JiT across model scales, training budgets, and image resolutions.
\end{itemize}

\section{Related Work}
\label{sec:related}

\subsection{Diffusion Models and Flow Matching.} Diffusion models~\citep{Ho2020, Song2021} have become a central framework for generative modeling by progressively transforming simple noise distributions into complex data distributions. Early formulations commonly adopt noise prediction ($\epsilon$-prediction)~\citep{Ho2020}, while subsequent work explores alternative parameterizations such as velocity ($v$) prediction~\citep{salimans2022progressive}, which can lead to different optimization behavior and training dynamics. Continuous-time formulations further connect diffusion processes with differential-equation-based generative modeling, providing a unified perspective on stochastic and deterministic sampling procedures~\citep{Song2021}. More recently, Flow Matching~\citep{albergo2023building,liu2022flow,lipman2023flowmatchinggenerativemodeling} provides a general framework for learning continuous vector fields along prescribed probability paths between a simple base distribution and the data distribution, enabling deterministic ODE-based generation without simulating the forward diffusion process during training. In parallel, generative backbones have increasingly shifted from convolutional U-Nets toward transformer architectures. Diffusion Transformers~\citep{Peebles2023} and Scalable Interpolant Transformers~\citep{ma2024sit} demonstrate that transformer-based architectures can effectively scale diffusion and flow-based generation by modeling long-range dependencies across visual tokens. Building on these developments, recent work has increasingly extended diffusion and Flow Matching directly to raw pixel space, where high-dimensional visual structure must be modeled without relying on a compressed latent representation. WiT focuses on the resulting finite-capacity optimization challenge and introduces semantic routing to better organize the learning of pixel-space transport directions.

\subsection{Generative Modeling in Pixel Space.} Direct generative modeling in pixel space has a long history across different paradigms. Autoregressive models such as PixelRNN~\citep{Oord2016a} and PixelCNN~\citep{Oord2016}, Generative Adversarial Networks~\citep{goodfellow2020generative,sauer2022stylegan}, and Normalizing Flows~\citep{dinh2017density,kingma2018glow} all model images directly in the observation space. Early diffusion models, including DDPM~\citep{Ho2020} and ADM~\citep{dhariwal2021diffusion}, likewise perform the generative process directly on pixels. Repeatedly processing high-dimensional spatial signals, however, makes pixel-space generation computationally demanding, particularly at high resolutions, which contributed to the widespread adoption of latent diffusion~\citep{Rombach2022}. While latent compression substantially reduces computational cost, reconstruction errors and information discarded by the tokenizer can ultimately constrain pixel-level fidelity.

Recent work has revisited end-to-end pixel-space generation from several complementary directions. Simple Diffusion~\citep{hoogeboom2023simple} and SiD2~\citep{hoogeboom2024simpler} demonstrate that carefully designed objectives and architectures can scale pixel diffusion to high resolutions, while PixelFlow~\citep{chen2025pixelflow} improves efficiency through cascaded flow modeling. Transformer-based approaches further investigate how to organize high-dimensional pixel prediction. JiT~\citep{Li_2026_CVPR} combines clean-image prediction with large-patch Transformers, PixNerd~\citep{wang2026pixnerd} introduces neural-field-based patch decoding, PixelDiT~\citep{Yu_2026_CVPR} combines global patch-level modeling with pixel-level refinement, DiP~\citep{Chen_2026_CVPR} decouples global generation from local detail recovery, and DeCo~\citep{ma2026deco} separates low- and high-frequency modeling. In parallel, self-supervised pretraining has emerged as another route for improving the representation quality and optimization of end-to-end pixel-space generators~\citep{lei_there_2026}.

Pixel-space modeling has also recently expanded toward unified multimodal understanding and generation. HiDream-O1-Image~\citep{cai2026hidream} integrates pixel-level image modeling with text and task conditioning in a unified transformer, while SenseNova-U1~\citep{diao2026sensenova} develops a native understanding-and-generation architecture without relying on a conventional visual encoder and VAE interface. Tuna-2~\citep{liu2026tuna} further explores raw pixel embeddings as a shared representation for multimodal understanding and generation, and Representation Forcing~\citep{wang2026representation} predicts intermediate visual representations to guide bottleneck-free pixel-space generation. These developments indicate that a central challenge in pixel-space modeling is how to organize learning effectively in the high-dimensional raw-pixel domain. WiT addresses this challenge from a complementary perspective by dynamically inferring compact semantic waypoints from the evolving noisy state and using them to spatially condition the Pixel Space Generator, while keeping the generative state and final prediction entirely in pixel space.

\subsection{Representation-Guided Optimization.} The representation space exposed to a generative model can strongly influence its optimization behavior. VA-VAE~\citep{yao2025reconstruction} studies the tension between reconstruction quality and generative optimization in latent diffusion and improves the generative representation by aligning autoencoder features with pretrained visual representations. REPA~\citep{repa} instead aligns intermediate diffusion features with pretrained semantic features during generative training. Subsequent works extend this direction in different ways. REPA-E~\citep{leng2025repa,repaet2i2025} couples representation alignment with end-to-end autoencoder tuning, while iREPA~\citep{singh2025matters} investigates which properties of pretrained representations are most important for generative alignment, emphasizing the role of spatial structure. Representation Autoencoders~\citep{zheng2026diffusion} directly adopt pretrained representation encoders as the basis of the generative latent space, and RAEv2~\citep{singh2026improved} further improves this framework while examining the complementarity between representation autoencoding and alignment-based optimization. PAE~\citep{yue2026matters} similarly studies the properties of diffusion-friendly latent manifolds and highlights the importance of semantic structure and spatial organization. Collectively, these studies show that the geometry and semantic organization of intermediate representations can substantially affect generative learning.

A related line of work makes semantic representations explicit components of the generation process rather than using them only as alignment targets. RCG~\citep{Li2023a} first generates self-supervised visual representations and subsequently conditions image generation on them. ReDi~\citep{kouzelis2026boosting} jointly models image representations and semantic features, exploiting their complementary information during generation. Representation Forcing~\citep{wang2026representation} similarly predicts intermediate visual representations to guide subsequent pixel-space synthesis in unified multimodal models. These methods illustrate the potential of explicitly exposing semantic representations during generation, although the representations are typically modeled as separate variables or stages of the generative procedure.

Recent studies have also examined representation-guided optimization directly within pixel-space and flow-based models. PixelREPA~\citep{shin2026representation} revisits representation alignment for JiT-style pixel-space Transformers and shows that directly transferring conventional alignment objectives from latent diffusion to raw-pixel generation is nontrivial. DeepFlow~\citep{shin2025deeply} improves flow-based generation by introducing supervision to intermediate network representations, while analyses of generative representation learning further show that architectural design choices can materially influence the quality of learned internal representations~\citep{zhang2024residual}. Together, these works reinforce a broader view in which effective generative optimization depends not only on the final prediction objective, but also on how useful intermediate structure is formed and exposed throughout the network.

WiT differs from prior representation-guided approaches by dynamically predicting a compact semantic waypoint from the current noisy pixel state rather than reshaping the latent space or imposing auxiliary feature alignment. The waypoint is recomputed as the pixel state evolves and conditions the Pixel Space Generator through Just-Pixel AdaLN, providing explicit semantic routing while keeping generation entirely in pixel space.

\section{Methodology}
\label{sec:method}

\begin{figure*}[t]
    \centering
    \includegraphics[width=\linewidth]{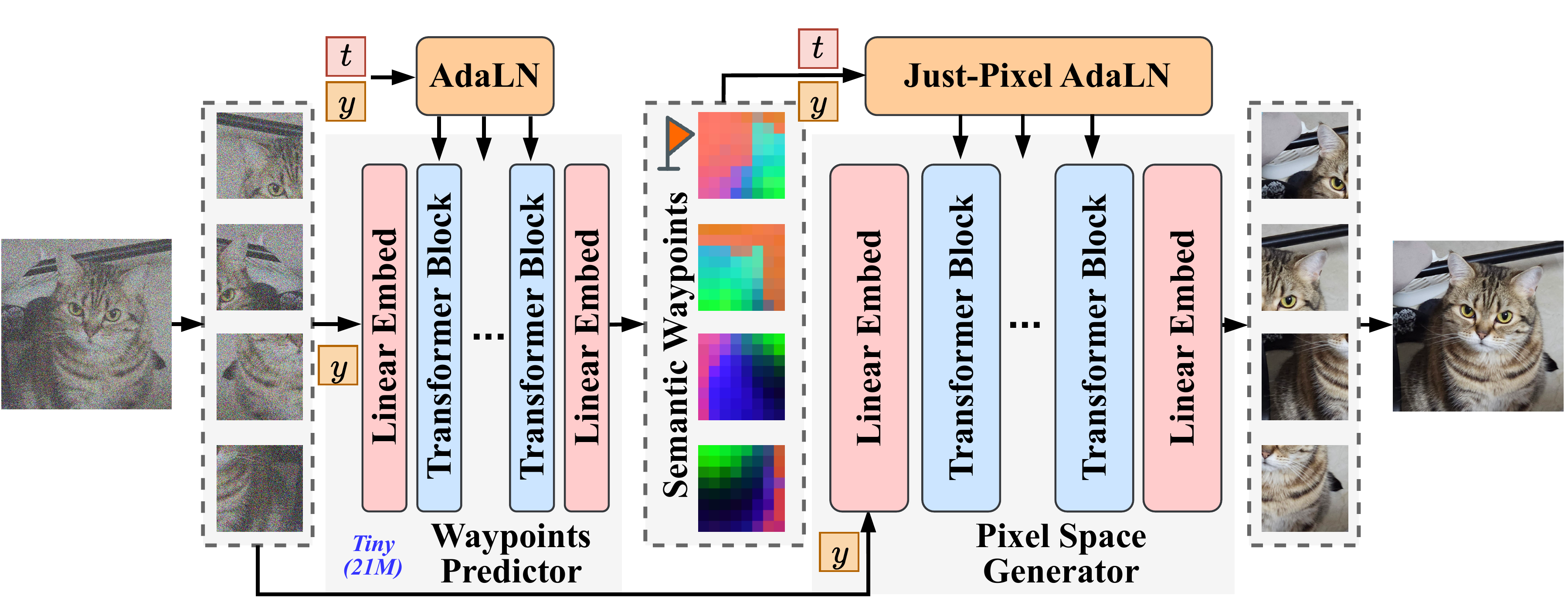}
    \caption{
    Overview of the WiT architecture.
    \textit{Left:} A lightweight Waypoints Predictor (21M parameters) dynamically predicts a compact semantic waypoint $\hat{s}_0$ from the current noisy pixel state $z_t$.
    \textit{Right:} The Pixel Space Generator directly predicts the clean image while using $\hat{s}_0$ as a spatial condition through the proposed Just-Pixel AdaLN mechanism.
    The generative state remains entirely in pixel space throughout inference.
    }
    \label{fig:arch}
\end{figure*}

In this section, we present the formulation and architecture of Waypoint Diffusion Transformers (WiT). We first review pixel-space Flow Matching and characterize the trajectory conflict considered in this work. We then construct compact semantic waypoints from pre-trained visual representations and introduce a lightweight Waypoints Predictor for dynamically predicting them from noisy pixel states. Finally, as illustrated in Figure~\ref{fig:arch}, we describe how the predicted waypoints condition the Pixel Space Generator through our Just-Pixel AdaLN mechanism, explicitly separating semantic representation prediction from high-fidelity pixel generation while keeping the generative state entirely in pixel space.

\subsection{Pixel-Space Flow Matching and Trajectory Conflict}
\label{subsec:prelim_conflict}

Following standard Flow Matching frameworks~\citep{liu2022flow,lipman2023flowmatchinggenerativemodeling}, let
$x \in \mathbb{R}^{H \times W \times 3}$ denote a clean target image and
$\epsilon \sim \mathcal{N}(0,\mathbf{I})$ denote standard Gaussian noise.
We consider the linear probability path
\begin{equation}
    z_t = t x + (1-t)\epsilon, \qquad t\in[0,1].
\label{eq:pixel_path}
\end{equation}
The corresponding path velocity is
\begin{equation}
    v^{\ast} = \frac{\mathrm{d}z_t}{\mathrm{d}t} = x-\epsilon = \frac{x-z_t}{1-t}.
\label{eq:pixel_velocity}
\end{equation}

Following JiT~\citep{Li_2026_CVPR}, we adopt $x$-prediction with a
$v$-space training objective. Specifically, rather than directly predicting
the velocity field, the Pixel Space Generator predicts the clean image
\begin{equation}
    \hat{x} = G_\theta(z_t,t,c),
\label{eq:x_prediction}
\end{equation}
where $c$ denotes the external condition. In our ImageNet class-conditional
setting, $c=y$ is the class label. The clean-image prediction is then
reparameterized into velocity space as
\begin{equation}
    \hat{v} = \frac{\hat{x}-z_t}{1-t}.
\label{eq:jit_velocity}
\end{equation}
Accordingly, the ideal $v$-loss can be written as
\begin{equation}
\begin{aligned}
    \mathcal{L}_{v} &= \mathbb{E}_{x,\epsilon,t,c} \left[ \left\| \hat{v}-v^{\ast} \right\|_2^2 \right] \\
    &= \mathbb{E}_{x,\epsilon,t,c} \left[ \frac{1}{(1-t)^2} \left\| \hat{x}-x \right\|_2^2 \right].
\end{aligned}
\label{eq:v_loss}
\end{equation}
Thus, the network directly predicts clean pixels, while the loss is evaluated
in the corresponding velocity parameterization.

In practice, following JiT~\citep{Li_2026_CVPR}, we lower-bound the denominator
by a small constant $\tau_{\mathrm{eps}}$ for numerical stability near
$t=1$. The velocity target and prediction used for optimization are therefore
\begin{equation}
    v = \frac{x-z_t}{\max(1-t,\tau_{\mathrm{eps}})}, \ \hat{v} = \frac{\hat{x}-z_t}{\max(1-t, \tau_{\mathrm{eps}})},
\label{eq:clipped_velocity}
\end{equation}
and the implemented objective is
\begin{equation}
    \mathcal{L}_{v} = \mathbb{E} \left[ \left\| \hat{v} - v \right\|_2^2 \right].
\label{eq:implemented_v_loss}
\end{equation}

However, even with external semantic conditioning, directly learning the pixel-space vector field remains challenging because local proximity in raw RGB space does not necessarily reflect similarity in target-specific structure. Along the interpolation $z_t=tx+(1-t)\epsilon$, noise progressively obscures pose, layout, scale, and other structural cues, while the required transport direction remains determined by the clean target. Consequently, at intermediate noise levels, two states can be close in the noisy pixel space yet correspond to substantially different target structures and transport requirements. The external condition $c$ constrains high-level semantics but does not resolve this target-specific variation, leaving these different requirements locally entangled for a shared finite-capacity regressor.

In this work, we use \textit{trajectory conflict} to describe the resulting finite-capacity regression phenomenon. Let $s_0$ denote the semantic representation of a clean target, defined in the next subsection, and let $d(\cdot,\cdot)$ denote an appropriate distance measure in each respective space. Consider two samples $i$ and $j$ at the same integration stage and under the same condition. Their noisy states can be locally similar,
\begin{equation}
    c_i=c_j,\qquad d\left(z_t^{(i)},z_t^{(j)}\right)\ \text{is small},
\label{eq:local_noisy_states}
\end{equation}
while their target semantic structures remain substantially different,
\begin{equation}
    d\left(s_0^{(i)},s_0^{(j)}\right)\ \text{is large}.
\label{eq:semantic_disagreement}
\end{equation}
Such locally similar states can therefore be associated with substantially different target-specific transport directions,
\begin{equation}
    v_i=x_i-\epsilon_i,\qquad v_j=x_j-\epsilon_j,\qquad d\left(v_i,v_j\right)\ \text{is large}.
\label{eq:velocity_disagreement}
\end{equation}

When different transport requirements are locally mixed in the noisy-state space, a shared finite-capacity regressor must resolve competing learning signals within similar input neighborhoods. Under regression-based Flow Matching~\citep{liu2022flow, lipman2023flowmatchinggenerativemodeling}, this can induce gradient interference and make the learned vector field under-resolve differences between target-specific transport directions~\citep{tong2024improving}. We refer to this optimization difficulty as \textit{trajectory conflict}.

WiT addresses this problem with a compact semantic waypoint $s_0$ that captures spatial target structure and serves as a routing and preconditioning bias. A lightweight Waypoints Predictor maps the noisy pixel observation to this semantic representation:
\begin{equation}
    \hat{s}_0 = W_\psi(z_t,t,c).
\label{eq:waypoint_routing}
\end{equation}

The predicted waypoint reorganizes the representation exposed to the finite-capacity Pixel Space Generator. By providing a semantically structured routing and preconditioning bias, WiT makes target-specific structure easier to represent and optimize from noisy pixel observations. The predicted waypoint is then explicitly exposed to the Pixel Space Generator through spatially varying Just-Pixel AdaLN modulation. This architecture provides the finite-capacity pixel generator with a structured intermediate representation, rather than requiring it to recover semantic and spatial organization implicitly while simultaneously modeling high-dimensional pixel details. Explicit semantic routing provides a more structured representation that helps the model better organize competing transport requirements and learn more accurate target-specific directions. We empirically examine this hypothesis in Section~\ref{sec:appendix_conflict} through a controlled toy experiment, local semantic-neighborhood analysis, and state-matched gradient interference.

\begin{figure*}[t]
    \centering
    \includegraphics[width=0.9\linewidth]{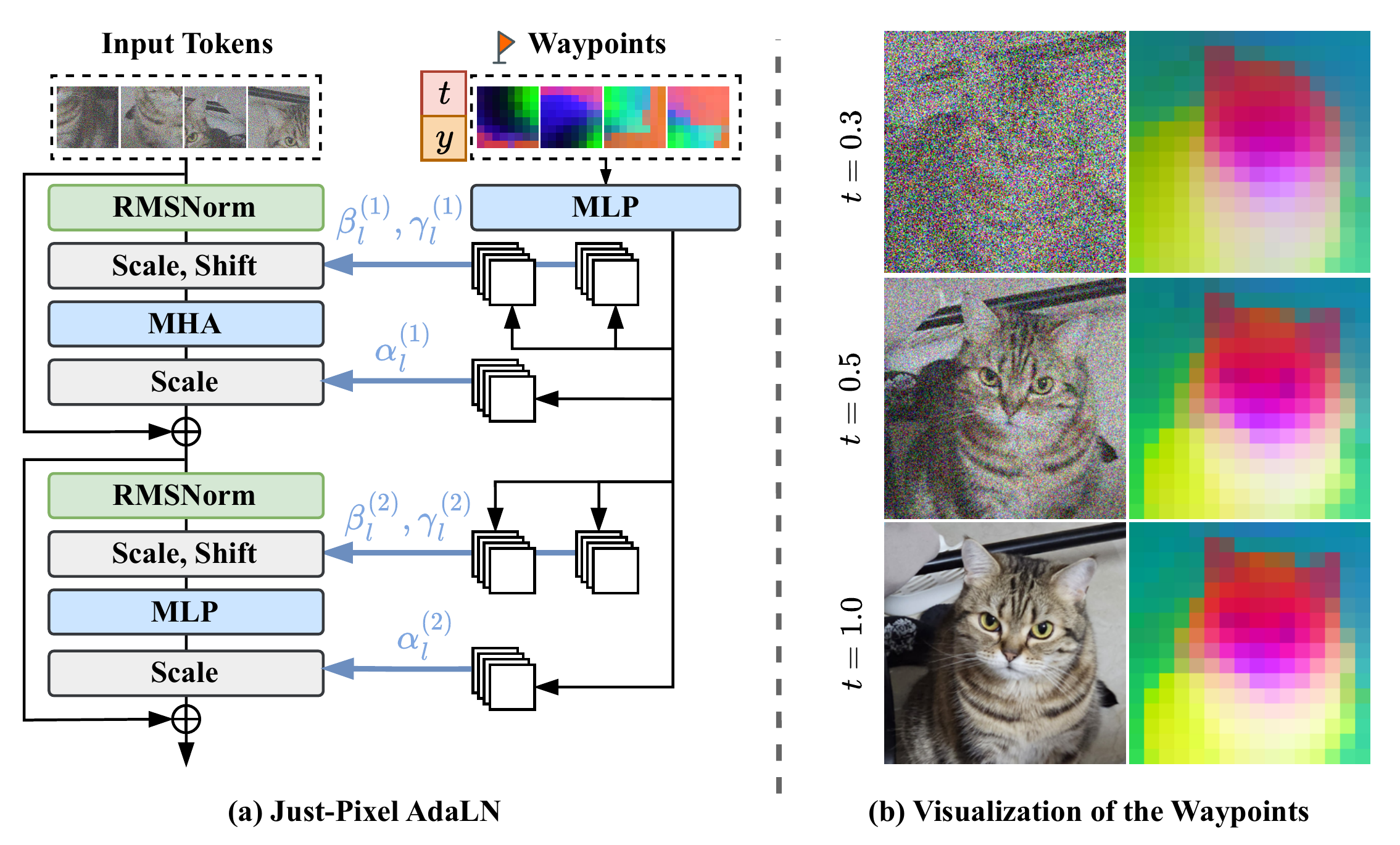}
    \caption{
    (a) Just-Pixel AdaLN. The predicted semantic waypoint provides spatially varying modulation to the Pixel Space Generator.
    (b) Visualization of the dynamically predicted semantic waypoints during inference.
    \textit{Left:} evolving noisy pixel states $z_t$ at different integration timesteps.
    \textit{Right:} the corresponding semantic waypoints $\hat{s}_0$ inferred by the lightweight Waypoints Predictor.
    }
    \label{fig:pca}
    \vspace{-10pt}
\end{figure*}

\subsection{Constructing Semantic Waypoints}

To provide explicit semantic structure for locally entangled pixel-space states, we leverage the semantically structured representation space of a frozen self-supervised vision model, specifically DINOv3~\citep{simeoni2025dinov3}. These representations provide dense structural cues that are not explicitly organized in raw RGB space.

For a target image $x$, we extract dense patch-wise visual features
\begin{equation}
    \phi(x) \in \mathbb{R}^{N\times D}.
\label{eq:dino_feature}
\end{equation}
Directly predicting the original high-dimensional representation would introduce substantial additional regression complexity. We therefore fit Principal Component Analysis (PCA) on the training distribution and retain its top $d$ principal directions. Let $U_d\in\mathbb{R}^{D\times d}$ denote the projection matrix and $\mu$ the dataset mean. The ground-truth semantic waypoint is defined as
\begin{equation}
    s_0 = \left( \phi(x)-\mu \right) U_d \in \mathbb{R}^{N\times d}, \qquad d=64.
\label{eq:pca}
\end{equation}

The semantic organization originates from the pre-trained DINOv3 representation, while PCA serves only as a compact linear projection that retains the dominant directions of feature variation. This substantially reduces the dimensionality of the waypoint prediction target while preserving dense spatial information required by the downstream Pixel Space Generator.

\paragraph{Lightweight Waypoints Predictor}

We introduce a lightweight transformer $W_\psi$ that predicts the clean semantic waypoint directly from the noisy pixel observation:
\begin{equation}
    \hat{s}_0 = W_\psi(z_t,t,y), \qquad z_t = t x+(1-t)\epsilon_{\mathrm{img}}.
\label{eq:waypoint_prediction}
\end{equation}
The timestep $t$ and class label $y$ are incorporated through the standard conditioning mechanism of the transformer.

Since our Pixel Space Generator follows the JiT~\citep{Li_2026_CVPR} $x$-prediction formulation with velocity-space supervision, we adopt an analogous training formulation for the Waypoints Predictor to maintain a consistent prediction and supervision scheme across the two stages. Specifically, the Waypoints Predictor directly predicts the clean semantic waypoint $\hat{s}_0$, while its supervision is constructed through the corresponding velocity parameterization in the semantic representation space.

To derive the corresponding timestep-dependent weighting, we introduce an auxiliary linear interpolation in the semantic representation space:
\begin{equation}
    z_{\mathrm{sem},t} = t s_0 + (1-t)\epsilon_{\mathrm{sem}}, \qquad \epsilon_{\mathrm{sem}} \sim \mathcal{N}(0, \mathbf{I}).
\label{eq:semantic_path}
\end{equation}
Under this interpolation, the semantic target and predicted velocities are
\begin{equation}
    v_{\mathrm{sem}} = \frac{s_0-z_{\mathrm{sem},t}}{1-t}, \qquad \hat{v}_{\mathrm{sem}} = \frac{\hat{s}_0 z_{\mathrm{sem},t}}{1-t}.
\label{eq:semantic_velocity}
\end{equation}

Following the same numerical stabilization used for pixel-space training, we lower-bound the denominator by $\tau_{\mathrm{eps}}$ in practice. The semantic training objective is therefore implemented as
\begin{equation}
\begin{aligned}
    \mathcal{L}_{\mathrm{sem}} &= \left\| \frac{ \hat{s}_0-z_{\mathrm{sem},t} }{ \max(1-t,\tau_{\mathrm{eps}}) } - \frac{ s_0-z_{\mathrm{sem},t} }{ \max(1-t,\tau_{\mathrm{eps}}) } \right\|_2^2 \\
    &= \frac{ \left\| \hat{s}_0-s_0 \right\|_2^2 }{ \max(1-t,\tau_{\mathrm{eps}})^2 }.
\end{aligned}
\label{eq:semantic_loss}
\end{equation}

Thus, the construction of $z_{\mathrm{sem},t}$ is introduced only to derive the timestep-dependent weighting in $\mathcal{L}_{\mathrm{sem}}$, which takes the $(1-t)^{-2}$ form before numerical clipping, rather than to define a semantic-space generative process.

During inference, the generative state remains entirely in pixel space. At each ODE evaluation, the current noisy pixel state $z_t$, together with the timestep and external condition, is fed into $W_\psi$ to predict the corresponding semantic waypoint $\hat{s}_0$. This waypoint is used only as a structured conditioning signal for the Pixel Space Generator and is recomputed as the pixel state evolves, rather than being propagated as a separate semantic state. In this way, WiT introduces dynamic semantic guidance throughout the pixel-space trajectory without requiring an additional semantic-space integration process. Because the waypoint target is compact ($d=64$), $W_\psi$ can be implemented with only 21M parameters, making it lightweight relative to the primary Pixel Space Generator.

\subsection{Semantic-Pixel Decoupled Architecture}

WiT structures the finite-capacity prediction problem through a decoupled architecture rather than decomposing the actual generative trajectory into separate flows. As shown in Figure~\ref{fig:arch}, it consists of a lightweight Waypoints Predictor $W_\psi$ and a primary Pixel Space Generator $G_\theta$. The former predicts an explicit semantic representation, while the latter retains responsibility for directly predicting clean RGB pixels.

\paragraph{Pixel Space Generator via Just-Pixel AdaLN}

Once the semantic waypoint $\hat{s}_0$ is predicted, it is provided to the primary Pixel Space Generator $G_\theta$ as a dense spatial condition. Unlike conventional global conditioning, which applies the same time-class representation to all spatial tokens, Just-Pixel AdaLN introduces spatially varying semantic modulation while preserving the native pixel-token sequence. As illustrated in Figure~\ref{fig:pca}(a), we combine the global time-class embedding with the projected semantic waypoint to form the spatial conditioning signal used throughout the Pixel Space Generator:
\begin{equation}
    c_{\mathrm{s}} = e(t,y) + \operatorname{Proj}(\hat{s}_0),
\label{eq:spatial_condition}
\end{equation}
where the embedding $e(t,y)$ is broadcast across spatial positions and $\operatorname{Proj}(\cdot)$ maps each 64-dimensional semantic token into the transformer hidden dimension $D_h$.

For the $l$-th transformer block, given hidden tokens $h_l\in\mathbb{R}^{N\times D_{\mathrm{h}}}$, the spatial condition is projected into six modulation parameters:
\begin{equation}
    \left( \gamma_l^{(1)}, \beta_l^{(1)}, \alpha_l^{(1)}, \gamma_l^{(2)}, \beta_l^{(2)}, \alpha_l^{(2)} \right) = \operatorname{MLP}_l(c_{\mathrm{s}}).
\label{eq:modulation}
\end{equation}

Following the AdaLN-Zero design~\citep{Peebles2023}, the spatial modulation is applied to the attention and MLP branches as
\begin{equation}
    \tilde{h}_l = h_l + \alpha_l^{(1)} \odot \mathcal{A} \left( (1+\gamma_l^{(1)})\odot\mathcal{N}(h_l) +\beta_l^{(1)} \right),
\label{eq:adaln_attn}
\end{equation}
and
\begin{equation}
    h_{l+1} = \tilde{h}_l + \alpha_l^{(2)} \odot \mathcal{M} \left( (1+\gamma_l^{(2)})\odot\mathcal{N}(\tilde{h}_l) +\beta_l^{(2)} \right),
\label{eq:adaln_mlp}
\end{equation}
where $\mathcal{A}$, $\mathcal{M}$, and $\mathcal{N}$ denote self-attention,
MLP, and RMSNorm~\citep{zhang2019root}, respectively.

By explicitly separating semantic waypoint prediction from high-dimensional pixel prediction, Just-Pixel AdaLN provides the Pixel Space Generator with spatially resolved semantic structure throughout the network without appending additional semantic tokens to the pixel sequence.

The Pixel Space Generator directly predicts the clean image $\hat{x}$ and is trained with the same $x$-prediction with $v$-loss formulation introduced above:
\begin{equation}
    \mathcal{L}_{\mathrm{img}} = \left\| \hat{v} - v \right\|_2^2,
\label{eq:limg}
\end{equation}
where $\hat{v}$ and $v$ are computed according to
Eq.~\ref{eq:clipped_velocity}.

\begin{algorithm}[t]
\caption{Training Procedure of WiT}
\label{alg:training}
\begin{algorithmic}[1]

\REQUIRE Dataset $\mathcal{D}$ and pretrained encoder $\phi$
\REQUIRE PCA projection $U_d$ and dataset mean $\mu$
\REQUIRE Waypoints Predictor $W_\psi$
\REQUIRE Pixel Space Generator $G_\theta$

\STATE \textbf{Stage 1: Train the Waypoints Predictor}

\WHILE{$W_\psi$ has not converged}

    \STATE Sample $(x,y)\sim\mathcal{D}$ and $t\sim p(t)$

    \STATE Apply class dropout to obtain $\tilde{y}$

    \STATE Sample
    $\epsilon_{\mathrm{img}}
    \sim\mathcal{N}(0,\mathbf{I})$

    \STATE
    $s_0\gets(\phi(x)-\mu)U_d$
    \COMMENT{Ground-truth waypoint}

    \STATE
    $z_t\gets tx+(1-t)\epsilon_{\mathrm{img}}$
    \COMMENT{Noisy pixel state}

    \STATE
    $\hat{s}_0\gets W_\psi(z_t,t,\tilde{y})$
    \COMMENT{Predict clean waypoint}

    \STATE
    $\mathcal{L}_{\mathrm{sem}} \gets \dfrac{ \|\hat{s}_0-s_0\|_2^2 }{ \max(1-t,\tau_{\mathrm{eps}})^2 }$

    \STATE Update $\psi$ using $\mathcal{L}_{\mathrm{sem}}$

\ENDWHILE

\STATE \textbf{Stage 2: Train the Pixel Space Generator}
\STATE Freeze the trained Waypoints Predictor $W_\psi$

\WHILE{$G_\theta$ has not converged}

    \STATE Sample $(x,y)\sim\mathcal{D}$ and $t\sim p(t)$

    \STATE Apply class dropout to obtain $\tilde{y}$

    \STATE Sample
    $\epsilon_{\mathrm{img}}\sim\mathcal{N}(0,\mathbf{I})$

    \STATE
    $z_t\gets tx+(1-t)\epsilon_{\mathrm{img}}$
    \COMMENT{Noisy pixel state}

    \STATE
    $\hat{s}_0\gets W_{\psi^\star}(z_t,t,\tilde{y})$
    \COMMENT{Predict semantic waypoint}

    \STATE
    $\hat{x}\gets G_\theta(z_t,t,\tilde{y},\hat{s}_0)$
    \COMMENT{Predict clean image}

    \STATE
    $\hat{v} \gets (\hat{x}-z_t)/ \max(1-t,\tau_{\mathrm{eps}})$

    \STATE
    $v \gets (x-z_t)/ \max(1-t,\tau_{\mathrm{eps}})$

    \STATE
    $\mathcal{L}_{\mathrm{img}} \gets \|\hat{v}-v\|_2^2$

    \STATE Update $\theta$ using $\mathcal{L}_{\mathrm{img}}$

\ENDWHILE

\end{algorithmic}
\end{algorithm}
\begin{algorithm}[t]
\caption{Inference Procedure of WiT}
\label{alg:inference}
\begin{algorithmic}[1]

\REQUIRE Frozen Waypoints Predictor $W_\psi$
\REQUIRE Pixel Space Generator $G_\theta$ with $L$ blocks
\REQUIRE Target class $y$ and integration steps $K$

\STATE
Sample initial pixel state
$z_{t_0}\sim\mathcal{N}(0,\mathbf{I})$

\STATE
Define integration schedule
$0=t_0<t_1<\cdots<t_K=1$

\STATE
$\mathcal{A},\mathcal{M},\mathcal{N}$ denote attention, MLP, and RMSNorm

\STATE
Integrate only $z_t$; recompute $\hat{s}_0$ at each ODE evaluation

\FOR{$k=0,\ldots,K-1$}

    \STATE \textbf{Dynamic Semantic Waypoint Prediction}

    \STATE
    $\hat{s}_0 \gets W_\psi(z_{t_k},t_k,y)$
    \COMMENT{Predict clean waypoint}

    \STATE \textbf{Spatial Condition Construction}

    \STATE
    $c_{\mathrm{s}} \gets e(t_k,y) + \operatorname{Proj}(\hat{s}_0)$
    \COMMENT{Spatial semantic condition}

    \STATE
    $h_1 \gets \operatorname{PatchEmbed}(z_{t_k})$
    \COMMENT{Pixel tokens}

    \STATE \textbf{Just-Pixel AdaLN Modulation}

    \FOR{$l=1,\ldots,L$}

        \STATE
        $\bigl( \gamma_l^{(1,2)}, \beta_l^{(1,2)}, \alpha_l^{(1,2)} \bigr) \gets \operatorname{Linear}_l(c_{\mathrm{s}})$

        \STATE
        $u_l^{(1)} \gets (1+\gamma_l^{(1)}) \odot \mathcal{N}(h_l) + \beta_l^{(1)}$

        \STATE
        $\tilde{h}_l \gets h_l + \alpha_l^{(1)} \odot \mathcal{A}(u_l^{(1)})$
        \COMMENT{Attention branch}

        \STATE
        $u_l^{(2)} \gets (1+\gamma_l^{(2)}) \odot  \mathcal{N}(\tilde{h}_l) + \beta_l^{(2)}$

        \STATE
        $h_{l+1} \gets \tilde{h}_l + \alpha_l^{(2)} \odot \mathcal{M}(u_l^{(2)})$
        \COMMENT{MLP branch}

    \ENDFOR

    \STATE \textbf{Clean-Image Prediction}

    \STATE
    $\hat{x} \gets \operatorname{LinearOut}(h_{L+1})$
    \COMMENT{Predict clean RGB image}

    \STATE \textbf{Pixel-Space Vector Field Estimation}

    \STATE
    $\hat{v}_{t_k} \gets (\hat{x}-z_{t_k})/ \max(1-t_k,\tau_{\mathrm{eps}})$

    \STATE \textbf{Pixel-State Update}
    
    \STATE
    Update $z_{t_k}\rightarrow z_{t_{k+1}}$
    using the Heun solver
    
\ENDFOR

\RETURN Generated image $z_{t_K}\approx x$

\end{algorithmic}
\vspace{-2.8pt}
\end{algorithm}

As summarized in Algorithm~\ref{alg:training}, WiT adopts a decoupled two-stage training strategy. The Waypoints Predictor $W_\psi$ is first trained to predict compact semantic waypoints from noisy pixel observations using the JiT-style clean-target prediction with velocity-space supervision. The trained Waypoints Predictor is subsequently frozen and used to provide spatially varying semantic conditions for the Pixel Space Generator $G_\theta$.

During inference, as summarized in Algorithm~\ref{alg:inference}, generation starts from Gaussian noise in pixel space. At each ODE evaluation, $W_\psi$ dynamically predicts a semantic waypoint from the current pixel state $z_t$. The waypoint is projected and combined with the global time-class embedding to construct the spatial condition $c_s$, which modulates the Pixel Space Generator through Just-Pixel AdaLN. The resulting clean-image prediction is converted into the pixel-space velocity field used by the ODE solver. Importantly, the semantic waypoint itself is never treated as an independently integrated generative state; instead, it is dynamically recomputed as a structured conditioning representation throughout the pixel-space trajectory, while $z_t$ remains the only state evolved by the ODE solver.

\section{Experimental Validation}
\label{sec:exp}

\subsection{Experimental Setup}
We conduct image generation experiments on ImageNet 2012~\citep{deng2009imagenet} at both $256\times256$ and $512\times512$ resolutions, and report Fr\'{e}chet Inception Distance~\citep{heusel2017gans} (FID-50K) and Inception Score (IS)~\citep{Salimans2016}. Unless otherwise specified, our WiT models and direct JiT baselines are evaluated with the 50-step Heun solver following JiT~\citep{Li_2026_CVPR}, while results for other generative models are taken from their corresponding publications.
\begin{table*}[t]
\caption{Configuration of Experiments for WiT.}
\label{tab:hyperparameters}
\centering
\begin{tabular}{lccccc}
\toprule
 & \textbf{WiT-B/16}
 & \textbf{WiT-L/16}
 & \textbf{WiT-XL/16}
 & \textbf{WiT-H/16}
 & \textbf{WiT-B/32} \\
\midrule
\multicolumn{6}{l}{\textbf{Architecture}} \\
\midrule
depth                  & 12  & 24   & 28   & 32   & 12  \\
hidden dim             & 768 & 1024 & 1152 & 1280 & 768 \\
heads                  & 12  & 16   & 16   & 16   & 12  \\
image size             & 256 & 256  & 256  & 256  & 512 \\
patch size             & 16  & 16   & 16   & 16   & 32  \\
bottleneck             & 128 & 128  & 128  & 256  & 128 \\
projection dropout     & 0.0 & 0.2  & 0.2  & 0.3  & 0.0 \\
attention dropout      & \multicolumn{5}{c}{0.0} \\
\midrule
\multicolumn{6}{l}{\textbf{Training}} \\
\midrule
epochs
  & \multicolumn{5}{c}{200 (ablation), 600} \\
warmup epochs
  & \multicolumn{5}{c}{5} \\
optimizer
  & \multicolumn{5}{c}{AdamW, $\beta_1=0.9, \beta_2=0.95$} \\
global batch size
  & \multicolumn{5}{c}{1024} \\
learning rate
  & \multicolumn{5}{c}{$5 \times 10^{-5}$} \\
learning rate schedule
  & \multicolumn{5}{c}{constant} \\
weight decay
  & \multicolumn{5}{c}{0} \\
ema decay
  & \multicolumn{5}{c}{$\{0.9999, 0.9996, 0.9998\}$} \\
time sampler
  & \multicolumn{5}{c}{
    $\operatorname{logit}(t) \sim
    \mathcal{N}(\mu,\sigma^2),\ \mu=-0.8,\ \sigma=0.8$
  } \\
noise scale
  & 1.0 & 1.0 & 1.0 & 1.0 & 2.0 \\
clip of $(1-t)$ in division
  & \multicolumn{5}{c}{0.05} \\
class token drop (for CFG)
  & \multicolumn{5}{c}{0.1} \\
\midrule
\multicolumn{6}{l}{\textbf{Sampling}} \\
\midrule
ODE solver
  & \multicolumn{5}{c}{Heun} \\
ODE steps
  & \multicolumn{5}{c}{50} \\
time steps
  & \multicolumn{5}{c}{linear in $[0.0,1.0]$} \\
CFG scale sweep range
  & \multicolumn{5}{c}{$[1.0,4.0]$} \\
CFG interval~\citep{kynkaanniemi2024applying}
  & \multicolumn{5}{c}{$[0.1,1.0]$} \\
\bottomrule
\end{tabular}%
\end{table*}

The Waypoints Predictor $W_\psi$ uses a lightweight ViT-S/16 architecture with 21M parameters. By default, semantic waypoint targets are constructed from frozen DINOv3-S/16~\citep{simeoni2025dinov3} features and reduced to $d=64$ dimensions using PCA. The PCA projection is estimated from 50,000 randomly sampled ImageNet training images. The primary Pixel Space Generator $G_\theta$ follows the corresponding JiT~\citep{Li_2026_CVPR} configurations at the Base, Large, Extra-Large, and Huge scales, allowing direct comparisons with the corresponding JiT generator backbones. The Waypoints Predictor is first trained for 600 epochs using the semantic objective described in Section~\ref{sec:method}. After training, its Exponential Moving Average (EMA)~\citep{tarvainen2017mean} weights are frozen and used to provide semantic waypoints during Pixel Space Generator training. The Pixel Space Generator is trained for up to 600 epochs. We use AdamW~\citep{loshchilov2018decoupled} with $\beta_1=0.9$ and $\beta_2=0.95$, a batch size of 1024, a base learning rate of $5\times10^{-5}$, a constant learning-rate schedule, and a 5-epoch linear warmup. Detailed configurations are summarized in Table~\ref{tab:hyperparameters}. Across all experiments, we use the same Waypoints Predictor trained for 600 epochs and keep its parameters fixed during subsequent Pixel Space Generator training and inference.

Following JiT~\citep{Li_2026_CVPR}, we sample the training timestep $t$ from a logit-normal distribution with $\mu=-0.8$ and $\sigma=0.8$. For numerical stability near $t=1$, the denominator $(1-t)$ in the velocity-space parameterization is lower-bounded by $0.05$. We use a class-token dropout rate of $0.1$ during training to enable Classifier-Free Guidance (CFG)~\citep{ho2022classifier}. During sampling, we employ truncated CFG following~\citep{kynkaanniemi2024applying} and evaluate its effect across different model capacities and training durations below.

\begin{table*}[t]
\caption{Comprehensive comparison of class-conditional ImageNet $256\times 256$.}
\label{tab:main-jit}
\centering
\setlength{\tabcolsep}{19pt}
\begin{tabular}{llccc}
\toprule
Method & Params & Epochs & IS $\uparrow$ & FID-50K $\downarrow$ \\
\midrule
\multicolumn{5}{l}{\textit{Latent-space Diffusion Models}} \\
DiT-XL/2~\citep{Peebles2023} & 675+49M & - & 278.2 & 2.27 \\
SiT-XL/2~\citep{ma2024sit} & 675+49M &  - & 277.5 & 2.06 \\
REPA (SiT-XL/2)~\citep{repa} & 675+49M & - & 305.7 & 1.42 \\
LightningDiT-XL/2~\citep{yao2025reconstruction} & 675+49M & - & 295.3 & 1.35 \\
DDT-XL/2~\citep{wang2026ddt} & 675+49M & - & 310.6 & 1.26 \\
RAE (DiT$^{\text{DH}}$-XL/2)~\citep{zheng2026diffusion} & 839+415M & - & 262.6 & \textbf{1.13} \\
\midrule
\multicolumn{5}{l}{\textit{Pixel-space Models (Non-diffusion)}} \\
JetFormer~\citep{tschannen2024jetformer} & 2.8B &  - & - & 6.64 \\
FractalMAR-H~\citep{li2025fractal} & 848M & - &  348.9 & 6.15 \\
\midrule
\multicolumn{5}{l}{\textit{Pixel-space Diffusion Models}} \\
ADM-G~\citep{dhariwal2021diffusion} & 554M &  - & 186.7 & 4.59 \\
RIN~\citep{jabri2022scalable} & 410M &  - & 182.0 & 3.42 \\
SiD (UViT/2)~\citep{hoogeboom2023simple} & 2B &  - & 256.3 & 2.44 \\
PixelFlow (XL/4)~\citep{chen2025pixelflow} & 677M &  - & 282.1 & 1.98 \\
PixNerd (XL/16)~\citep{wang2026pixnerd} & 700M &  - & 297.0 & 2.15 \\
\midrule
\multicolumn{5}{l}{\textit{Direct Baselines \& Ours}} \\
JiT-B/16~\citep{Li_2026_CVPR} & 131M &  200 & - & 4.37 \\
\textbf{WiT-B/16 (Ours)} & 131M+21M &  200 & {270.7} & \textbf{3.34} \\
\addlinespace
JiT-B/16~\citep{Li_2026_CVPR} & 131M &  600 & 275.1 & 3.66 \\
\textbf{WiT-B/16 (Ours)} & 131M+21M & 600 & {280.2} & \textbf{3.03} \\
\cdashline{1-5}
JiT-L/16~\citep{Li_2026_CVPR} & 459M &  200 & - & 2.79 \\
LF-DIT-L/16~\citep{baade2026latent} & 465M & 200 & - & 2.48 \\
\textbf{WiT-L/16 (Ours)} & 459M+21M & 200 & {276.6} & \textbf{2.28} \\
\addlinespace
JiT-L/16~\citep{Li_2026_CVPR} & 459M &  600 & 298.5 & 2.36 \\
\textbf{WiT-L/16 (Ours)} & 459M+21M &  600 & {293.3} & \textbf{2.02} \\
\cdashline{1-5}
\textbf{WiT-XL/16 (Ours)} & 676M+21M & 200 & 288.9 & 2.16 \\
\textbf{WiT-XL/16 (Ours)} & 676M+21M & 600 & {301.0} & \textbf{1.89} \\
\cdashline{1-5}
JiT-H/16~\citep{Li_2026_CVPR} & 953M &  600 & 303.4 & 1.86 \\
JiT-G/16~\citep{Li_2026_CVPR} & 2B &  600 & 292.6 & {1.82} \\
\textbf{WiT-H/16 (Ours)} & 953M+21M & 600 & {302.3} & \textbf{1.79} \\
\bottomrule
\end{tabular}
\vspace{-4pt}
\end{table*}

\subsection{Main Results}

\paragraph{Quantitative Results}
We compare WiT against a comprehensive set of representative generative models, including latent-space diffusion models (\emph{e.g.}, DiT~\citep{Peebles2023} and SiT~\citep{ma2024sit}), pixel-space non-diffusion models (\emph{e.g.}, JetFormer~\citep{tschannen2024jetformer}), and pure pixel-space diffusion models (\emph{e.g.}, PixelFlow~\citep{chen2025pixelflow}, PixNerd~\citep{wang2026pixnerd}, and our direct baseline JiT~\citep{Li_2026_CVPR}). As shown in Table~\ref{tab:main-jit}, WiT consistently improves FID over the corresponding JiT baselines across backbone scales and generator training durations, demonstrating clear gains in generation quality.

\begin{figure*}[t]
    \centering
    \includegraphics[width=\linewidth]{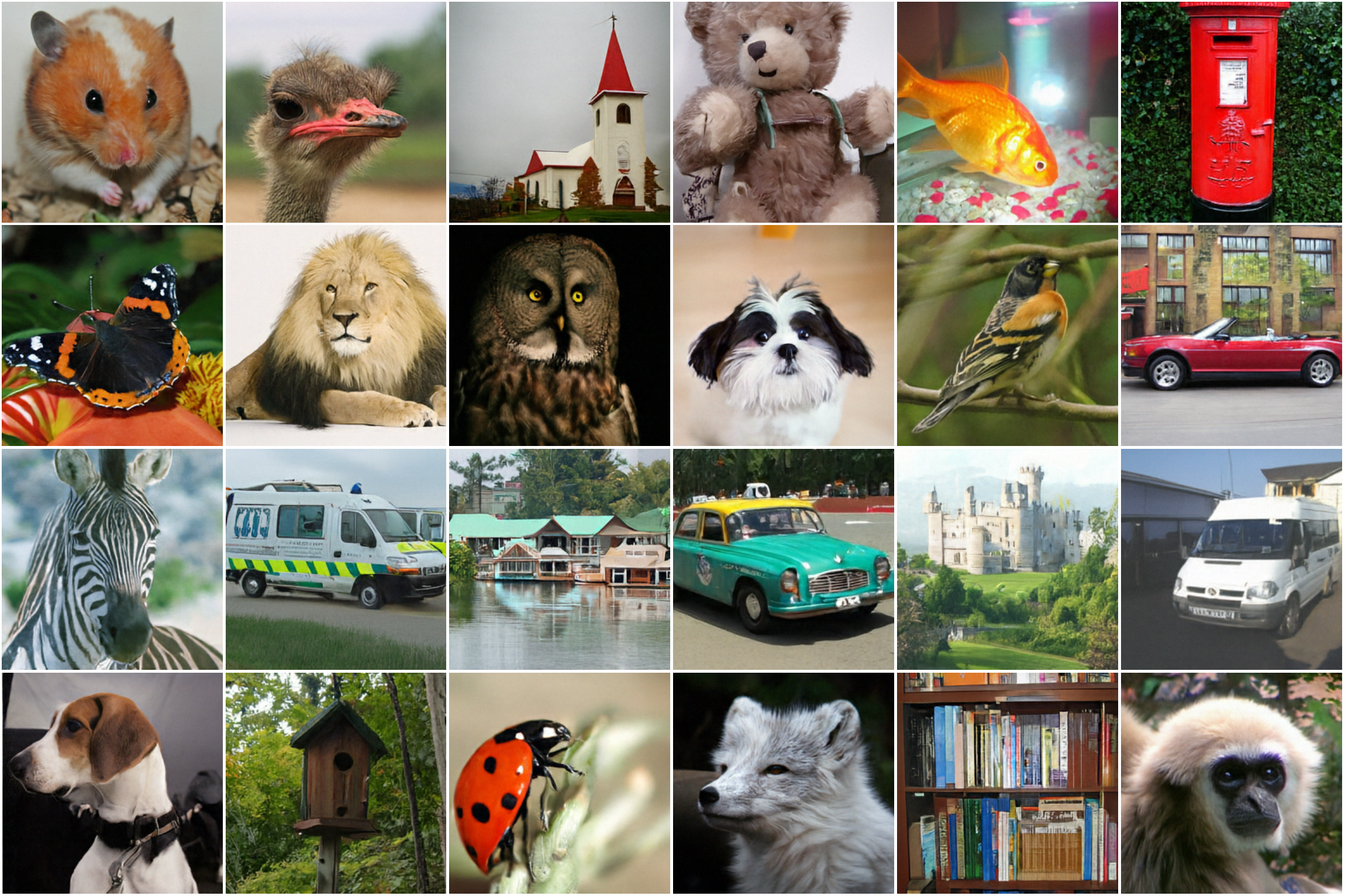}
    \caption{Qualitative Results of WiT-L/16 on ImageNet $256 \times 256$~\citep{deng2009imagenet}.}
    \label{fig:qualitative}
\end{figure*}

With the B/16 configuration, WiT achieves an FID of 3.34 after 200 epochs, substantially improving over JiT-B/16 (FID 4.37). With the B/16 configuration, WiT achieves an FID of 3.34 after 200 epochs, substantially improving over JiT-B/16 at 4.37. Extending training to 600 epochs further improves the FID to 3.03, compared with 3.66 for JiT-B/16.

At the L/16 scale, WiT-L/16 achieves an FID of 2.28 after 200 epochs, improving over JiT-L/16 at 2.79 and LF-DiT-L/16 at 2.48 after the same number of generator training epochs. When trained for 600 epochs, WiT-L/16 further reduces the FID to 2.02, compared with 2.36 for the matched JiT-L/16 baseline. This result also outperforms DiT-XL/2~\citep{Peebles2023}, which reports an FID of 2.27.

The improvement remains consistent as the model capacity increases. WiT-XL/16 reaches an FID of 2.16 after 200 epochs and 1.89 after 600 epochs, demonstrating that the proposed semantic routing mechanism continues to benefit larger pixel-space generators. At the Huge scale, WiT-H/16 achieves an FID of 1.79 and an IS of 302.3 after 600 epochs,
improving over the matched JiT-H/16 baseline with an FID of 1.86. Notably, WiT-H/16 also achieves a lower FID than the substantially larger JiT-G/16 model, which reports an FID of 1.82 with 2B parameters. Overall, these results show that the gains from semantic waypoint routing are consistent across model scales and generator training durations, while preserving
generation in pixel space.

\begin{figure}[t]
    \centering
    \includegraphics[width=\linewidth]{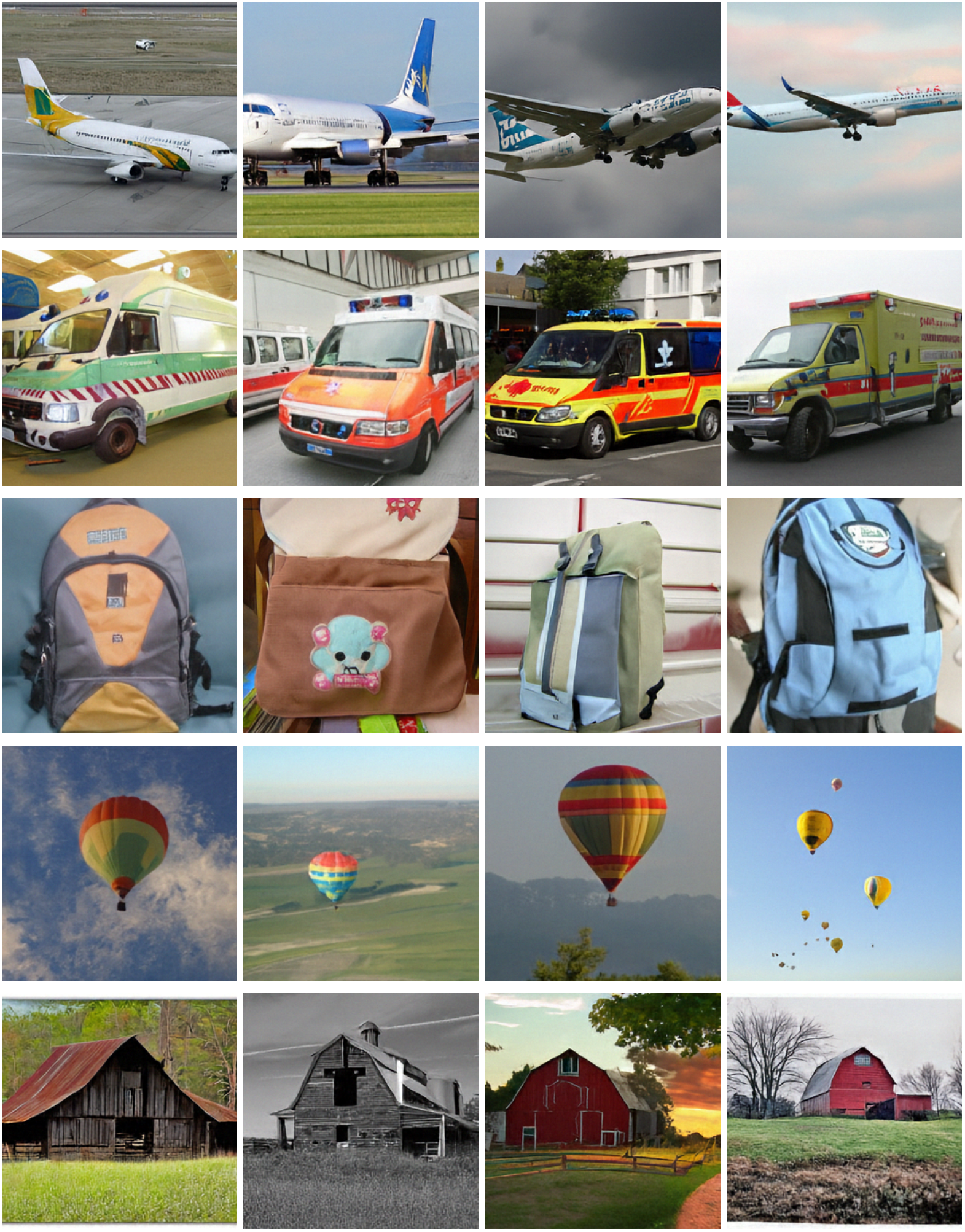}
    \caption{Visual Results of WiT-XL/16 on ImageNet $256 \times 256$~\citep{deng2009imagenet}.}
    \label{fig:xl_visual}
\end{figure}

\begin{figure}[t]
    \centering
    \includegraphics[width=\linewidth]{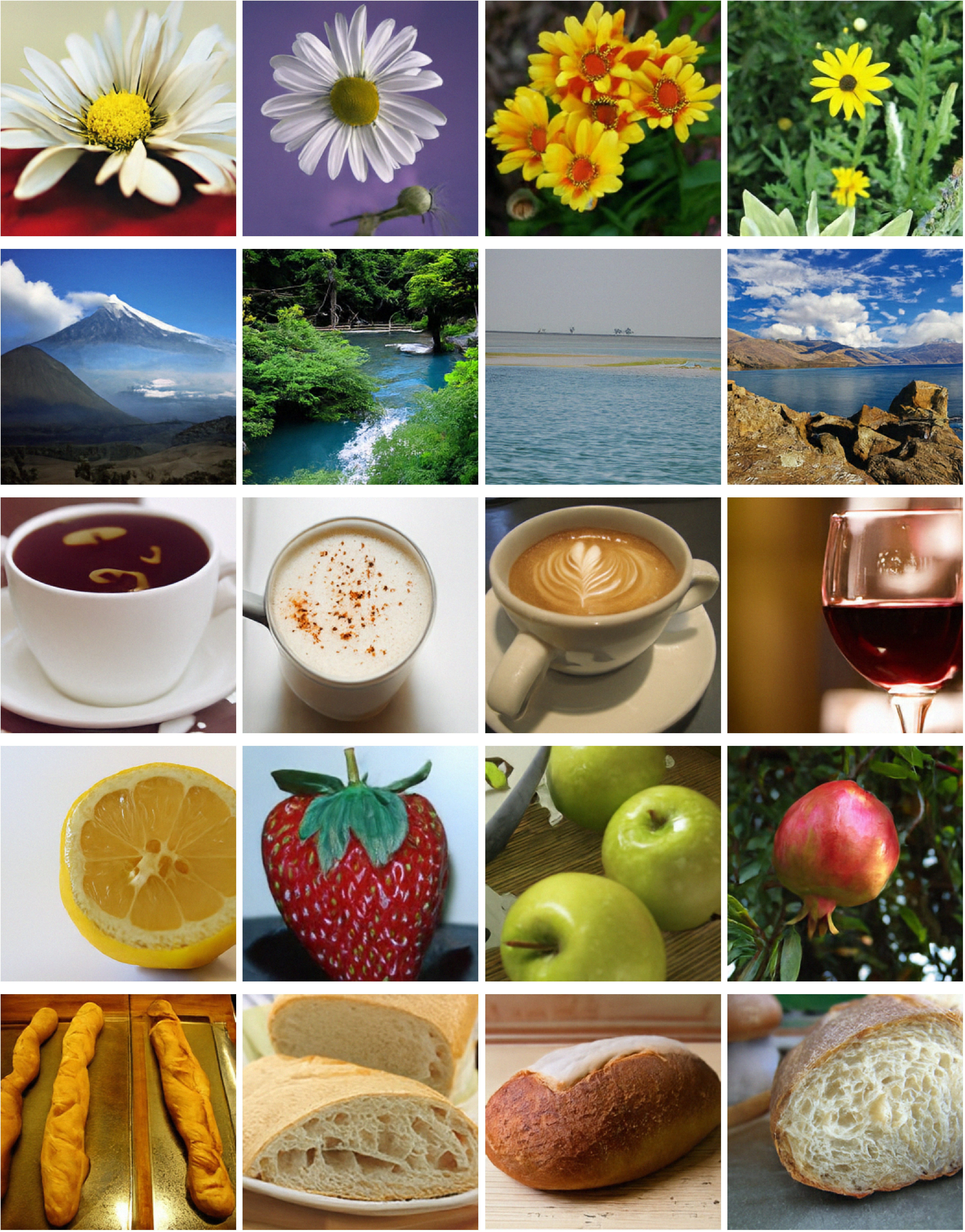}
    \caption{Visual Results of WiT-H/16 on ImageNet $256 \times 256$~\citep{deng2009imagenet}.}
    \label{fig:h_visual}
\end{figure}

\paragraph{Qualitative Results} Figure~\ref{fig:qualitative} presents representative ImageNet $256\times256$ samples generated by WiT-L/16, complementing the quantitative improvements reported above with qualitative evidence. Across diverse object categories and scene layouts, WiT produces coherent global structures while preserving fine-grained local details. For example, animals and structured scenes such as the lion and castle exhibit plausible proportions, object layouts, and spatial relationships, suggesting that the dynamically predicted semantic waypoints provide useful target-relevant guidance throughout generation. At the same time, because the generative state remains entirely in pixel space, the model retains detailed local appearance, including fine feather patterns, fur textures, and intricate butterfly wings. These observations are consistent with the intended role of semantic routing. To further assess visual quality at larger model scales, we provide additional generated samples from WiT-XL/16 and WiT-H/16 in Figures~\ref{fig:xl_visual} and~\ref{fig:h_visual}, respectively.

\begin{table}[t]
\caption{
Scaling results on ImageNet $512\times512$.
}
\vspace{-2pt}
\label{tab:imagenet512}
\centering
\setlength{\tabcolsep}{9pt}
\begin{tabular}{lcc}
\toprule
Method
& 200 Epochs
& 600 Epochs \\
\midrule
JiT-B/32
& 4.64
& 4.02 \\
\textbf{WiT-B/32 (Ours)}
& \textbf{3.89}
& \textbf{3.38} \\
\bottomrule
\end{tabular}
\vspace{-10pt}
\end{table}
\paragraph{Scaling to ImageNet $512\times512$} We further evaluate WiT at a higher image resolution to examine whether the benefit of semantic waypoint routing extends beyond the default $256\times256$ setting. Specifically, we train WiT-B/32 on ImageNet $512\times512$ following the training protocol of JiT~\citep{Li_2026_CVPR}. As shown in Table~\ref{tab:imagenet512}, WiT-B/32 achieves an FID of 3.89 after 200 epochs, substantially improving over the 4.64 obtained by JiT-B/32. With 600 epochs of training, WiT-B/32 further reduces the FID to 3.38, compared with 4.02 for JiT. These results indicate that semantic waypoint routing remains effective when pixel-space generation is scaled to higher image resolution.

\subsection{Ablation Studies}

We conduct a series of ablations to examine the main design choices of WiT and better understand the contribution of each component. Specifically, we study waypoint dimensionality, semantic injection strategies, alternative representation-guided approaches, and different pretrained visual representations. We also analyze the effect of CFG across model capacities and training durations to characterize the sampling behavior of WiT under different settings.

\paragraph{Waypoint PCA Dimension $d$} We vary the number of retained PCA components to study the effect of waypoint dimensionality. As shown in Table~\ref{tab:ablation}, reducing $d$ from $64$ to $32$ degrades the FID from 3.34 to 5.11, while increasing it to $128$ results in an FID of 4.12. These results suggest a trade-off between retaining sufficient structural variation in the semantic waypoint and keeping the prediction target compact enough for the lightweight Waypoints Predictor to learn reliably from noisy pixel observations. An overly compressed waypoint may lose useful semantic structure, whereas a higher-dimensional target can introduce unnecessary regression difficulty. We therefore use $d=64$ throughout all the experiments.

\begin{table*}[t]
\caption{
Comparison with alternative representation-guided strategies on
ImageNet $256\times256$. All methods use the JiT-B/16 backbone and are
trained for 200 epochs under the same setting.
}
\label{tab:representation_guidance}
\centering
\setlength{\tabcolsep}{13pt}
\begin{tabular}{lccccc}
\toprule
Method
& JiT-B/16
& + REPA
& + RCG
& + PixelREPA
& WiT-B/16 (Ours) \\
\midrule
FID $\downarrow$
& 4.37
& 5.14
& 4.96
& 4.00
& \textbf{3.34} \\
\bottomrule
\end{tabular}
\vspace{-6pt}
\end{table*}
\begin{figure*}[t]
    \centering
    \begin{subfigure}[b]{0.326\textwidth}
        \centering
        \includegraphics[width=\textwidth]{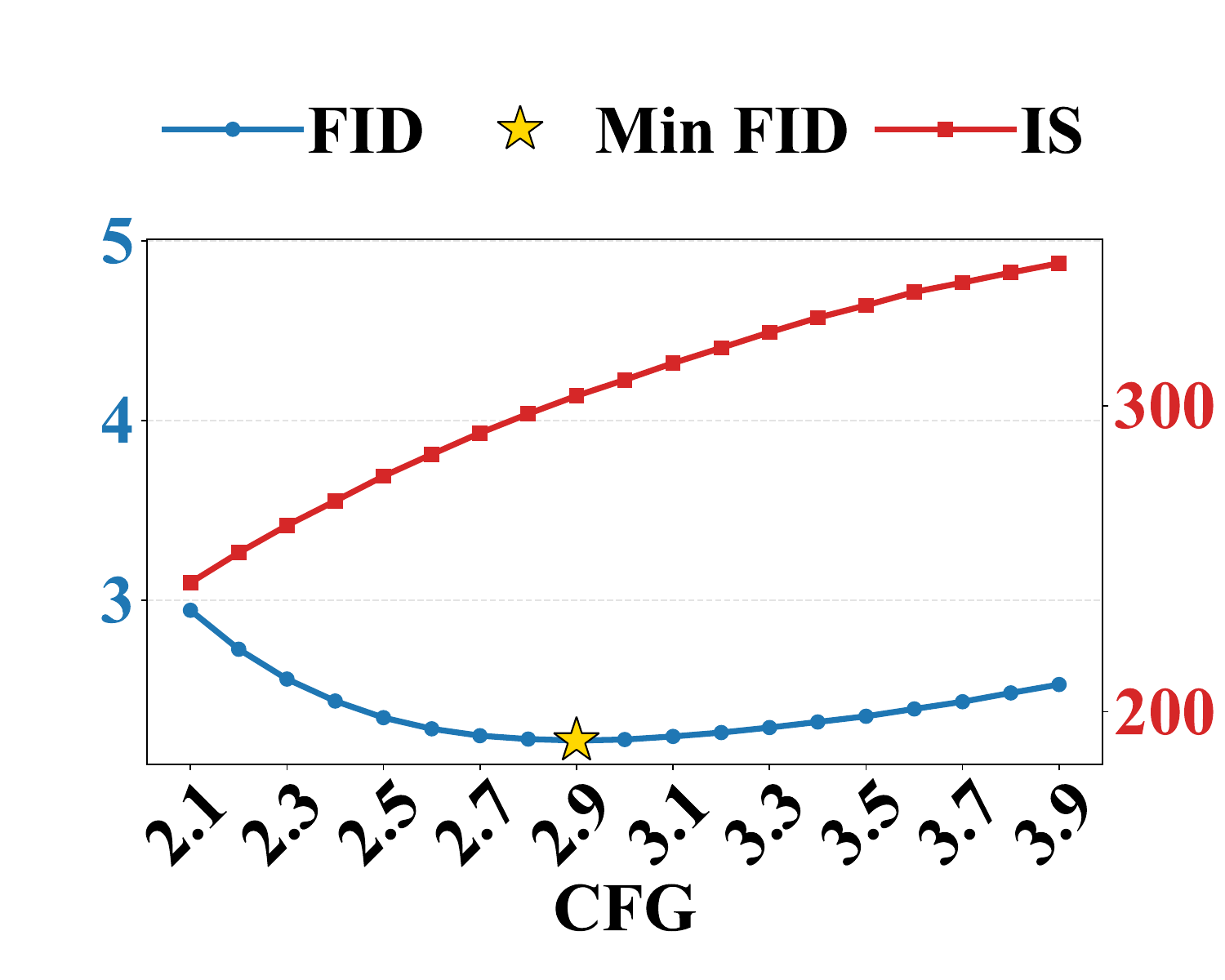}
        \caption{WiT-L, 600 epoch}
        \label{fig:wit_l_600}
    \end{subfigure}
    \hfill
    \begin{subfigure}[b]{0.326\textwidth}
        \centering
        \includegraphics[width=\textwidth]{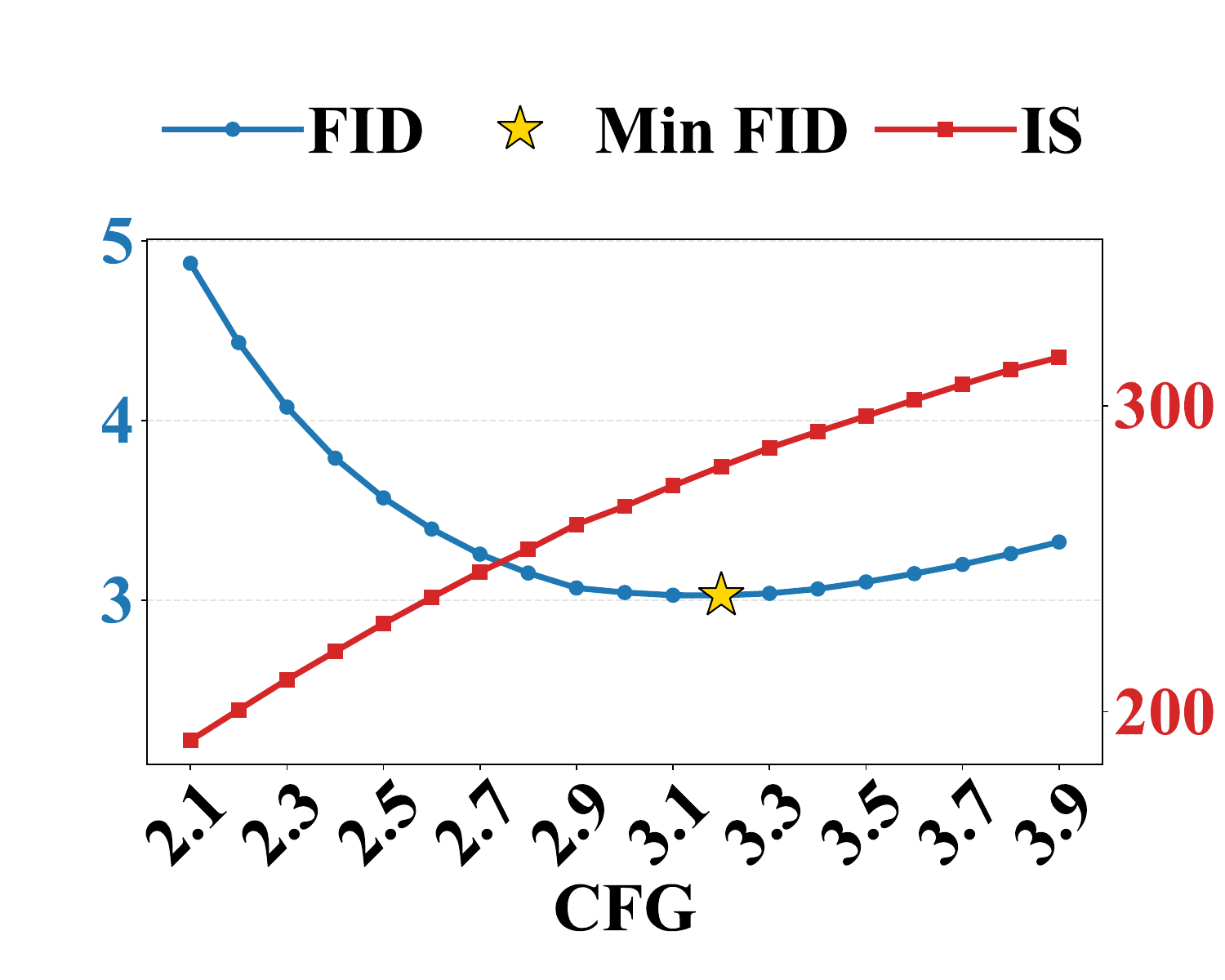}
        \caption{WiT-B, 600 epoch}
        \label{fig:wit_b_600}
    \end{subfigure}
    \hfill
    \begin{subfigure}[b]{0.326\textwidth}
        \centering
        \includegraphics[width=\textwidth]{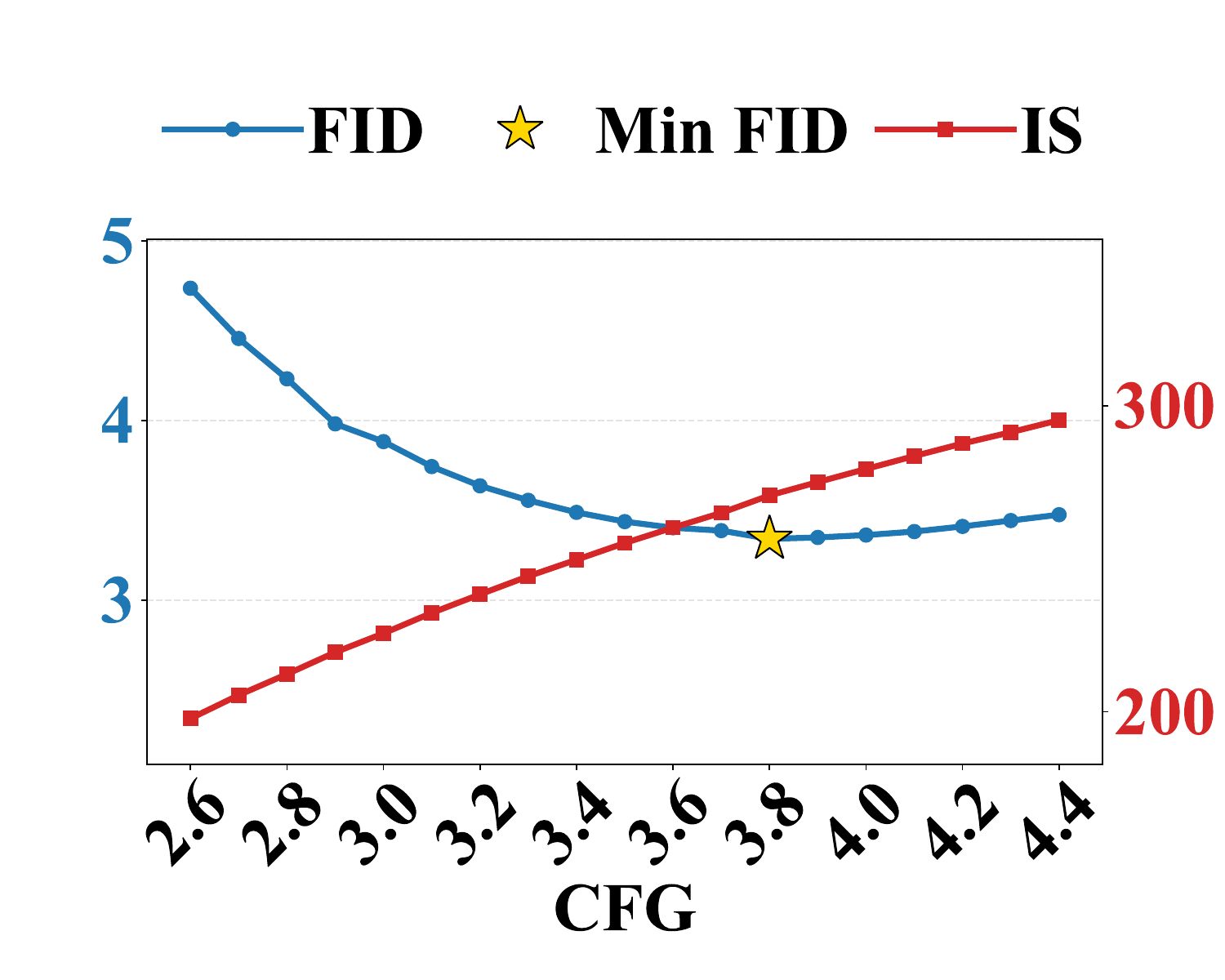}
        \caption{WiT-B, 200 epoch}
        \label{fig:wit_b_200}
    \end{subfigure}
    \caption{The impact of CFG on FID and IS. The gold star indicates the minimum FID.}
    \label{fig:cfg_comparison}
    \vspace{-6pt}
\end{figure*}

\paragraph{Semantic Injection Strategy} We compare three mechanisms for incorporating the predicted semantic waypoint into the Pixel Space Generator. Channel Concat concatenates $\hat{s}_0$ with noisy pixel patches along the channel dimension, In-context Concat appends semantic tokens to the transformer sequence, and Just-Pixel AdaLN applies spatially varying affine modulation without extending the native pixel-token sequence. As shown in Table~\ref{tab:ablation}, Channel Concat and In-context Concat achieve FIDs of 3.93 and 3.63, while Just-Pixel AdaLN performs best with an FID of 3.34 and an IS of 270.73. This indicates that spatially aligned modulation is more effective than direct feature or token concatenation.

\paragraph{Comparison with Representation-Guided Alternatives}
Since WiT leverages pretrained visual representations to construct semantic waypoints, we further compare it with alternative strategies for introducing representation guidance into the same JiT-B/16 backbone. As shown in Table~\ref{tab:representation_guidance}, the vanilla JiT-B/16 baseline obtains an FID of 4.37 after 200 epochs. Directly incorporating REPA~\citep{repa} and RCG~\citep{Li2023a} results in FIDs of 5.14 and 4.96, respectively, while PixelREPA~\citep{shin2026representation} improves the baseline to 4.00. In comparison, WiT-B/16 achieves the best FID of 3.34. For all variants, we retain the same JiT-B/16 backbone and modify only the corresponding representation-guidance mechanism.

These results indicate that the gain of WiT is not simply attributable to the presence of pretrained representation guidance. Under the evaluated setting, explicitly predicting a target-relevant semantic waypoint from the current noisy state and using it as a spatially varying routing condition is more effective than the alternative representation-guided strategies.

\begin{table*}[t]
\caption{
Effect of different pretrained visual representations for constructing
semantic waypoints. All waypoint-based variants use the B/16 Pixel Space
Generator and are trained for 600 epochs.
}
\label{tab:pretrained_representation}
\centering
\setlength{\tabcolsep}{11pt}
\begin{tabular}{lccccc}
\toprule
Representation
& JiT
& MoCoV3-S/16
& MAE-B/16
& DINOv2-S/14
& DINOv3-S/16 \\
\midrule
FID $\downarrow$
& 3.66
& 3.11
& 3.32
& 3.36
& \textbf{3.03} \\
\bottomrule
\end{tabular}
\end{table*}

\paragraph{Choice of Pretrained Representation} We investigate whether WiT depends on a specific pretrained visual representation by constructing semantic waypoints from several self-supervised encoders while keeping the Pixel Space Generator and training configuration unchanged. As shown in Table~\ref{tab:pretrained_representation}, all evaluated representations improve over the JiT-B/16 baseline with an FID of 3.66. MoCoV3-S/16~\citep{chen2021empirical}, MAE-B/16~\citep{he2022masked}, and DINOv2-S/14~\citep{oquab2023dinov2} achieve FIDs of 3.11, 3.32, and 3.36, respectively, while DINOv3-S/16~\citep{simeoni2025dinov3} performs best with an FID of 3.03. For comparison, we use encoders with a $16\times16$ patch size whenever possible. MoCoV3-S/16 and DINOv3-S/16 are naturally aligned with the B/16 Pixel Space Generator, while the DINOv2-S/14 feature map is spatially aligned to the $16\times16$ waypoint grid before PCA projection. For MAE, we use the available B/16 configuration. All evaluated pretrained representations improve over the JiT baseline, while DINOv3-S/16 achieves the best FID. We therefore use DINOv3-S/16 as the default encoder.

\paragraph{CFG Scale} Finally, we study the effect of CFG across model capacities and training durations. Figure~\ref{fig:cfg_comparison} shows that increasing the CFG scale initially improves FID while increasing IS, whereas overly strong guidance eventually degrades FID. The minimum-FID scale shifts from 3.8 for WiT-B/16 at 200 epochs to 3.1 at 600 epochs and 2.9 for WiT-L/16 at 600 epochs. This suggests that larger or more fully trained WiT models achieve their best FID with milder guidance.

\section{Quantitative Analysis of Trajectory Conflict}
\label{sec:appendix_conflict}

Building upon the discussion in Section~\ref{subsec:prelim_conflict}, we further investigate trajectory conflict in pixel-space Flow Matching. In this work, we use \textit{trajectory conflict} to describe a finite-capacity regression phenomenon in which nearby noisy states at the same stage and under the same external condition can correspond to substantially different target structures and be associated with different transport corrections. When these locally entangled requirements cannot be sufficiently separated by a shared finite-capacity model, their learning signals can interfere, making target-specific transport directions more difficult to resolve.

The predicted waypoint $\hat{s}_0$ is inferred from the current noisy state and condition and is explicitly exposed to the Pixel Space Generator as a compact semantic representation. In this way, WiT provides a routing and preconditioning bias for finite-capacity learning. We examine this hypothesis through a controlled two-spiral experiment, local semantic-neighborhood analysis, and state-matched gradient interference.
\begin{table*}[t]
\caption{Ablation studies on WiT-B/16 (200 epochs). We investigate the impact of waypoint PCA dimension ($d$) and the architectural injection method.}
\label{tab:ablation}
\centering
\setlength{\tabcolsep}{20pt}
    \begin{tabular}{lcp{3cm}cc}
        \toprule
        Configuration & PCA ($d$) & Injection & IS $\uparrow$ & FID $\downarrow$ \\
        \midrule
        \multicolumn{5}{l}{\textit{Ablation on Waypoint PCA Dimension}} \\
        WiT-B/16 & 32 & Just-Pixel AdaLN & 210.40 & 5.11 \\
        WiT-B/16 & 128 & Just-Pixel AdaLN & 211.33  & 4.12 \\
        \midrule
        \multicolumn{5}{l}{\textit{Ablation on Injection Mechanism}} \\
        WiT-B/16 & 64 & Channel Concat & 221.19  & 3.93 \\
        WiT-B/16 & 64 & In-context Concat & 238.92  & 3.63 \\
        \midrule
        \textbf{WiT-B/16 (Ours)} & \textbf{64} & \textbf{Just-Pixel AdaLN} & \textbf{270.73} & \textbf{3.34} \\
        \bottomrule
    \end{tabular}
\end{table*}

\subsection{Toy Experiment}

We first examine semantic routing on a two-spiral dataset, where the target manifold and the deviation of generated samples from this manifold can be measured directly. The data are embedded into normalized observation spaces of different dimensionalities $d$. Under matched model capacity and optimization settings, we compare a jointly optimized baseline serving as a JiT baseline, a Random Route baseline using irrelevant random routing labels, and WiT with target-relevant semantic routing.

For $N$ generated samples $\{\hat{x}_i\}_{i=1}^{N}$ and a discretized target manifold $\{\mu_k\}$ in the normalized $d$-dimensional observation space, we measure the normalized nearest-manifold RMSE:
\begin{equation}
    \mathrm{nRMSE} = \sqrt{ \frac{1}{Nd} \sum_{i=1}^{N} \min_k \left\| \hat{x}_i-\mu_k \right\|_2^2 }.
\label{eq:toy_nrmse}
\end{equation}

\begin{figure*}[t]
    \centering
    \includegraphics[width=\linewidth]{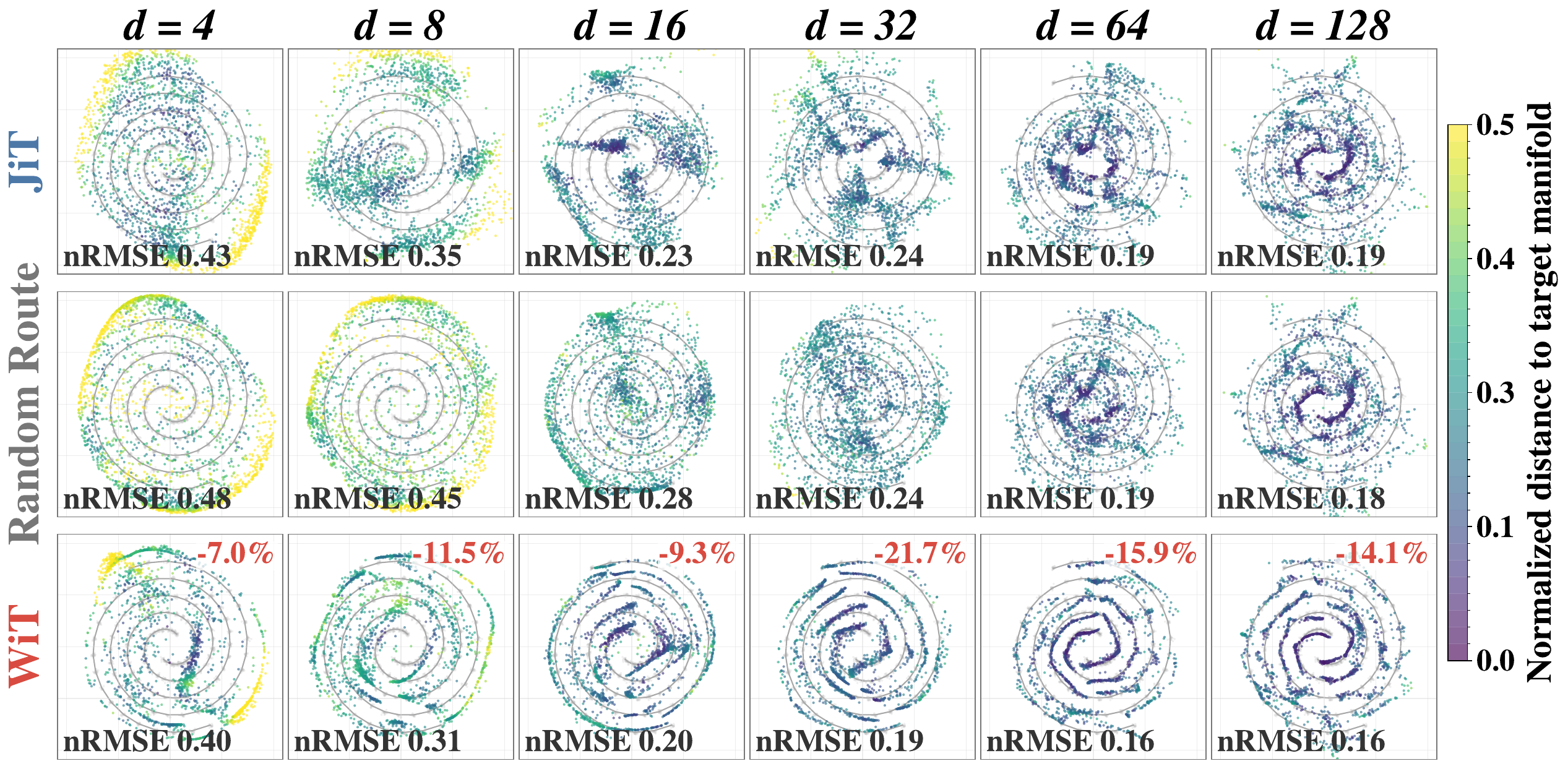}
    \caption{
    Toy experiment on two-spiral data embedded into observation spaces of different dimensionalities $d$.
    The samples are visualized using a two-dimensional projection, while the normalized nearest-manifold distances used for coloring and nRMSE are computed in the original $d$-dimensional observation space.
    We compare a jointly optimized JiT baseline, a parameter-matched Random Route baseline using irrelevant routing labels, and WiT with target-relevant semantic routing.
    Gray curves denote the target manifold, and lower nRMSE indicates better manifold recovery.
    Red percentages report the relative nRMSE reduction of WiT with respect to the JiT baseline.
    }
    \label{fig:toy_dimension}
\end{figure*}

As shown in Figure~\ref{fig:toy_dimension}, WiT achieves lower nRMSE than the jointly optimized baseline at every evaluated dimensionality. The relative reductions are $7.0\%$, $11.5\%$, $9.3\%$, $21.7\%$, $15.9\%$, and $14.1\%$ for $d=4,8,16,32,64,$ and $128$, respectively. In contrast, the Random Route baseline does not provide a comparable and consistently positive improvement. These results highlight the benefit of target-relevant semantic routing for manifold recovery.

\subsection{Local Semantic Entanglement of Noisy States}

We next examine local semantic entanglement in the ImageNet setting. If nearby noisy states under the same external condition are associated with substantially different semantic targets, a finite-capacity vector-field model must resolve different target-specific transport requirements within the same local region. We therefore measure the semantic dispersion of local neighborhoods while fixing the external condition.

For each anchor sample $i$, we construct a neighborhood in a query representation $q$ by including the anchor and its $k$ nearest neighbors:
\begin{equation}
    \mathcal{G}_i(q) = \{i\} \cup \mathcal{N}_k(i;q), \qquad m = |\mathcal{G}_i(q)|.
\label{eq:nlcv_group}
\end{equation}
All neighbors are restricted to the same external condition as the anchor. Let
\begin{equation}
    \bar{s}_i = \frac{1}{m} \sum_{j\in\mathcal{G}_i(q)} s_0^{(j)}
\label{eq:nlcv_local_mean}
\end{equation}
denote the mean ground-truth semantic waypoint within the local neighborhood. We compute its unbiased semantic variance as
\begin{equation}
    \widehat{V}_i(q) = \frac{1}{m-1} \sum_{j\in\mathcal{G}_i(q)} \left\| s_0^{(j)}-\bar{s}_i \right\|_2^2.
\label{eq:nlcv_local_variance}
\end{equation}

For each condition $c$, let $n_c$ denote the number of samples and
\begin{equation}
    \bar{s}_c = \frac{1}{n_c} \sum_{j:c_j=c} s_0^{(j)}
\label{eq:nlcv_class_mean}
\end{equation}
its semantic mean. The corresponding unbiased within-condition semantic variance is
\begin{equation}
    \widehat{V}_c = \frac{1}{n_c-1} \sum_{j:c_j=c} \left\| s_0^{(j)}-\bar{s}_c \right\|_2^2.
\label{eq:nlcv_class_variance}
\end{equation}
We define the Normalized Local Conditional Variance (NLCV) by macro-averaging the normalized local dispersion across conditions:
\begin{equation}
    \mathcal{D}_{\mathrm{NLCV}}(q,t) = \frac{1}{|\mathcal{C}|} \sum_{c\in\mathcal{C}} \frac{1}{n_c} \sum_{i:c_i=c} \frac{ \widehat{V}_i(q) }{ \widehat{V}_c+\epsilon }.
\label{eq:nlcv}
\end{equation}

For the noisy-state geometry, neighbor search is performed using the PCA-whitened patch-mean representation
\begin{equation}
    q_{\mathrm{raw}}=\operatorname{WhitenPCA}\left(\operatorname{Flatten}\left(\operatorname{PatchMean}(z_t)\right)\right),
\label{eq:raw_query_rep}
\end{equation}
while the waypoint geometry is evaluated using
\begin{equation}
    q_{\mathrm{sem}}=\operatorname{WhitenPCA}\left(\operatorname{MeanPool}(\hat{s}_0)\right).
\label{eq:semantic_query_rep}
\end{equation}
We use $k=4$ nearest neighbors, yielding local groups of $m=5$ samples including the anchor. Neighbors are restricted to the same ImageNet class and selected in a 64-dimensional PCA-whitened query space. The raw query is constructed from patch-mean noisy-state features, while the semantic query is obtained by mean-pooling the predicted waypoint tokens.

\begin{table}[t]
\caption{
Within-condition Normalized Local Conditional Variance (NLCV).
Lower values indicate that nearby samples correspond to more coherent ground-truth semantic endpoints.
}
\label{tab:nlcv}
\centering
\setlength{\tabcolsep}{5pt}
    \begin{tabular}{cccc}
        \toprule
        $t$
        & Raw State
        & Pred. Semantic
        & Reduction \\
        \midrule
        0.16 & 0.9713 & 0.7401 & 23.8\% \\
        0.31 & 0.9668 & 0.6578 & 32.0\% \\
        0.51 & 0.9636 & 0.6208 & 35.6\% \\
        \bottomrule
    \end{tabular}
\end{table}

As shown in Table~\ref{tab:nlcv}, neighborhoods constructed from the noisy-state representation remain strongly mixed in terms of their semantic endpoints even after the external condition is fixed. In contrast, neighborhoods formed in the predicted waypoint representation exhibit substantially lower semantic dispersion, with reductions ranging from $23.8\%$ to $35.6\%$ across the evaluated timesteps. These results show that the predicted waypoint representation forms local neighborhoods with more coherent target semantics. This provides a more structured representation for the downstream finite-capacity Pixel Space Generator.

\subsection{State-Matched Gradient Interference}

The local semantic analysis establishes that nearby noisy observations can correspond to substantially different target structures. We next examine whether this target disagreement translates into optimization interference, which constitutes the optimization-level manifestation of trajectory conflict.

We compute per-sample gradients with respect to the output-layer parameters shared by JiT and WiT, denoted by $\Theta_{\mathrm{out}}$. Let $\ell_i$ denote the image-generation training loss for sample $i$, and define
\begin{equation}
    g_i = \nabla_{\Theta_{\mathrm{out}}} \ell_i.
\label{eq:sample_gradients}
\end{equation}
For a sample pair $(i,j)$, the gradient agreement is measured by
\begin{equation}
    \mathcal{G}_{ij} = \frac{ g_i^{\top}g_j }{ \|g_i\|_2 \|g_j\|_2 +\epsilon}.
\label{eq:gradient_cosine}
\end{equation}

To isolate the effect of semantic target disagreement, we match sample pairs by external condition, timestep, and noisy-state distance, and then divide them into high- and low-disagreement groups according to their semantic target discrepancy. For model $M$, we define the high-minus-low gradient-cosine effect as
\begin{equation}
    \Delta\mathcal{G}_{M} = \mathbb{E} \left[ \mathcal{G}_{ij} \mid \mathrm{high},M \right] - \mathbb{E} \left[ \mathcal{G}_{ij} \mid \mathrm{low},M \right].
\label{eq:gradient_effect}
\end{equation}
A more negative value indicates that larger target disagreement induces stronger gradient opposition among otherwise state-matched samples.

We additionally measure the negative gradient mass
\begin{equation}
    \mathcal{M}_{ij}^{-} = \max \left( 0,-\mathcal{G}_{ij} \right),
\label{eq:negative_gradient_mass}
\end{equation}
and its corresponding high-minus-low effect
\begin{equation}
    \Delta\mathcal{M}_{M}^{-} = \mathbb{E} \left[ \mathcal{M}_{ij}^{-} \mid \mathrm{high},M \right] - \mathbb{E} \left[ \mathcal{M}_{ij}^{-} \mid \mathrm{low},M \right].
\label{eq:negative_gradient_mass_effect}
\end{equation}

Because weaker trajectory conflict corresponds to a less negative $\Delta\mathcal{G}$ and a smaller positive $\Delta\mathcal{M}^{-}$, the WiT mitigation quantities are computed as
\begin{equation}
\begin{aligned}
    \operatorname{Mitigation}_{\mathcal{G}} &= \Delta\mathcal{G}_{\mathrm{WiT}} - \Delta\mathcal{G}_{\mathrm{JiT}}, \\
    \operatorname{Mitigation}_{\mathcal{M}} &= \Delta\mathcal{M}_{\mathrm{JiT}}^{-} - \Delta\mathcal{M}_{\mathrm{WiT}}^{-}.
\end{aligned}
\label{eq:gradient_mitigation}
\end{equation}

As reported in Table~\ref{tab:gradient_conflict}, higher semantic target disagreement is associated with stronger output-layer gradient interference for both models. This disagreement-dependent effect is weaker in WiT, with a less negative high-minus-low gradient cosine and a smaller increase in negative gradient mass. These results provide optimization-level evidence for our trajectory-conflict interpretation. Among otherwise state-matched samples, greater target disagreement is associated with more conflicting learning signals, while semantic routing reduces this disagreement-dependent interference.

\subsection{Discussion}

Taken together, these analyses characterize trajectory conflict as a finite-capacity learning difficulty arising from locally mixed target-specific transport requirements. The NLCV analysis shows that local noisy-state neighborhoods can remain associated with substantially different semantic targets even under the same external condition, indicating that target-specific structure is not well separated in noisy pixel space. The state-matched gradient analysis further shows that larger semantic disagreement is associated with stronger output-layer gradient interference, while this effect is weaker in WiT. These observations provide complementary representation- and optimization-level evidence for trajectory conflict and support semantic waypoints as a structured routing representation for separating locally mixed target-specific requirements.

The toy experiment further supports this interpretation by showing that target-relevant semantic routing improves manifold recovery. More broadly, these results suggest that semantic waypoints reorganize local representation geometry and provide useful routing and preconditioning for pixel-space generation. By repeatedly exposing target-relevant semantic structure as the noisy state evolves, WiT helps the finite-capacity generator distinguish locally mixed transport requirements. These analyses provide complementary evidence for the proposed mechanism and help explain the benefit of semantic routing in pixel-space learning.

\begin{table*}[t]
\caption{State-matched analysis of disagreement-dependent output-layer gradient interference. Sample pairs are matched by condition, timestep, and noisy-state distance, and then grouped according to semantic target disagreement. Confidence intervals are estimated using 10,000 ImageNet-class cluster bootstrap samples. Relative mitigation is measured with respect to the corresponding JiT effect.}

\label{tab:gradient_conflict}
\centering
\setlength{\tabcolsep}{8pt}
    \begin{tabular}{lcccc}
        \toprule
        Metric
        & JiT
        & WiT
        & WiT Mitigation
        & Relative Mitigation \\
        \midrule
        High-minus-low gradient cosine
        & -0.15961
        & -0.13677
        & $0.02284$ $[0.00200,\,0.04391]$
        & \textbf{14.3\%} \\
        High-minus-low negative mass
        & 0.07434
        & 0.05833
        & $0.01601$ $[0.00451,\,0.02730]$
        & \textbf{21.5\%} \\
        \bottomrule
    \end{tabular}
\end{table*}


\section{Conclusion}
\label{sec:conclusion}

In this paper, we presented Waypoint Diffusion Transformers (WiT), a pixel-space generative framework that introduces explicit semantic routing to alleviate trajectory conflict in Flow Matching. WiT structures the finite-capacity prediction problem through dynamically inferred semantic waypoints, which condition the Pixel Space Generator via our Just-Pixel AdaLN mechanism while keeping the evolving generative state entirely in pixel space. Our analyses show that the predicted waypoints reorganize local semantic neighborhoods and reduce disagreement-dependent gradient interference, providing complementary geometric and optimization-level evidence for the proposed mechanism. The semantic waypoint is dynamically recomputed from the current noisy state throughout generation and serves as a structured routing condition. Experiments on ImageNet at both $256\times256$ and $512\times512$ resolutions demonstrate consistent improvements over the corresponding JiT baselines across model scales and training durations, with WiT-H/16 achieving an FID of 1.79 at $256\times256$. These results show that the benefit of semantic routing persists across different model capacities, training budgets, and image resolutions. Overall, our results suggest that dynamically routing pixel-space prediction through compact semantic waypoints provides an effective way to organize high-dimensional generative learning while preserving direct generation in the raw pixel domain.

\section*{Statements and Declarations}

The authors have no competing interests to declare that are relevant to the content of this article.

\section*{Data Availability}

The ImageNet 2012 dataset analyzed in this study is available from the ImageNet project at \url{https://www.image-net.org/}, subject to its access terms.

\section*{Acknowledgements}
We would like to thank Qiming Hu for the insightful discussions and feedback. The computational resources of this work was partially supported by TPU Research Cloud (TRC).

\bibliographystyle{spbasic}
\bibliography{sn-bibliography}

\end{document}